\documentclass{article}

\usepackage{arxiv}

\usepackage[utf8]{inputenc} 
\usepackage[T1]{fontenc}    
\usepackage{hyperref}       
\usepackage{url}            
\usepackage{booktabs}       
\usepackage{amsfonts}       
\usepackage{nicefrac}       
\usepackage{microtype}      
\usepackage{graphicx}
\usepackage{natbib}
\usepackage{doi}

\newcommand{\degre}{\ensuremath{^\circ}}
\usepackage{amsmath}  
\usepackage{tikz}
\usepackage{float}

\title{k-textures, a self-supervised hard clustering deep learning algorithm for satellite image segmentation}

\author{ \href{https://orcid.org/0000-0002-9623-1182}{\includegraphics[scale=0.06]{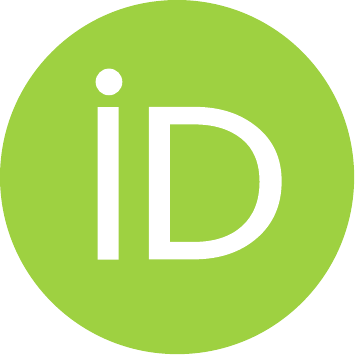}\hspace{1mm}Fabien H.~Wagner}\\
	Institute of the Environment and Sustainability,\\
	University of California, Los Angeles, CA 90095 USA;\\
	and	NASA-Jet Propulsion Laboratory\\
	California Institute of Technology, Pasadena, CA 91109, USA \\ 
	\texttt{wagner.h.fabien@gmail;fhwagner@ucla.edu} \\
 	\And
 	\href{https://orcid.org/0000-0002-7151-8697}{\includegraphics[scale=0.06]{orcid.pdf}\hspace{1mm}Ricardo Dalagnol} \\
	Institute of the Environment and Sustainability,\\
	University of California, Los Angeles, CA 90095 USA;\\
	and	NASA-Jet Propulsion Laboratory\\
	California Institute of Technology, Pasadena, CA 91109, USA \\  
 	\texttt{ricds@hotmail.com } \\
 	 	\And
 	\href{https://orcid.org/0000-0001-7966-2880}{\includegraphics[scale=0.06]{orcid.pdf}\hspace{1mm}Alber H.~Sánchez} \\
 	Earth Observation and Geoinformatics Division\\
 	National Institute for Space Research – INPE\\
 	São José dos Campos, SP, Brazil\\
 	\texttt{alber.ipia@inpe.br} \\
 	 	\And
 	\href{https://orcid.org/0000-0002-1817-360X}{\includegraphics[scale=0.06]{orcid.pdf}\hspace{1mm}Mayumi C.M.~Hirye} \\ 
 	Quapá Lab\\
 	Faculty of Architecture and Urbanism\\
 	University of São Paulo---USP\\
 	São Paulo, SP, Brazil\\
 	\texttt{mayhirye@hotmail.com} \\
 		\And
 	\href{https://orcid.org/0000-0002-8600-1949}{\includegraphics[scale=0.06]{orcid.pdf}\hspace{1mm}Samuel Favrichon} \\
	NASA-Jet Propulsion Laboratory\\
	California Institute of Technology,\\
	Pasadena, CA 91109, USA \\ 
 	\texttt{samuel.favrichon@jpl.nasa.gov} \\
 		\And
 	\href{https://orcid.org/0000-0002-2838-2878}{\includegraphics[scale=0.06]{orcid.pdf}\hspace{1mm} Jake H.~Lee} \\
	NASA-Jet Propulsion Laboratory\\
	California Institute of Technology,\\
	Pasadena, CA 91109, USA \\ 
 	\texttt{jake.h.lee@jpl.nasa.gov}\\
 	 		\And
 	\href{https://orcid.org/0000-0003-2371-8843}{\includegraphics[scale=0.06]{orcid.pdf}\hspace{1mm} Steffen Mauceri} \\
	NASA-Jet Propulsion Laboratory\\
	California Institute of Technology,\\
	Pasadena, CA 91109, USA \\ 
 	\texttt{steffen.mauceri@jpl.nasa.gov}\\
 	 		\And
  	\href{https://orcid.org/0000-0002-0526-4561}{\includegraphics[scale=0.06]{orcid.pdf}\hspace{1mm} Yan Yang} \\
 	NASA-Jet Propulsion Laboratory\\
 	California Institute of Technology,\\
 	Pasadena, CA 91109, USA \\ 
  	\texttt{yangyannn@gmail.com}\\
  	     	\And
 	\href{https://orcid.org/0000-0001-8524-4917}{\includegraphics[scale=0.06]{orcid.pdf}\hspace{1mm}Sassan Saatchi} \\
	Institute of the Environment and Sustainability,\\
	University of California, Los Angeles, CA 90095 USA;\\
	and	NASA-Jet Propulsion Laboratory\\
	California Institute of Technology, Pasadena, CA 91109, USA \\ 
 	\texttt{sasan.s.saatchi@jpl.nasa.gov}\\
}

\renewcommand{\shorttitle}{k-textures algorithm}

\begin{document}
\maketitle

\newpage

\begin{abstract}
Deep learning self-supervised algorithms that can segment an image in a fixed number of hard labels such as the k-means algorithm and relying only on deep learning techniques are still lacking. Here, we introduce the k-textures algorithm which provides self-supervised segmentation of a 4-band image (RGB-NIR) for a $k$ number of classes. An example of its application on high resolution Planet satellite imagery is given. Our algorithm shows that discrete search is feasible using convolutional neural networks (CNN) and gradient descent. The model detects $k$ hard clustering classes represented in the model as $k$ discrete binary masks and their associated $k$ independently generated textures, that combined are a simulation of the original image. The similarity loss is the mean squared error between the features of the original and the simulated image, both extracted from the penultimate convolutional block of Keras 'imagenet' pretrained VGG-16 model and a custom feature extractor made with Planet data. The main advances of the k-textures model are: first, the $k$ discrete binary masks are obtained inside the model using gradient descent. The model allows for the generation of discrete binary masks using a novel method using a hard sigmoid activation function. Second, it provides hard clustering classes -- each pixels has only one class. Finally, in comparison to k-means, where each pixel is considered independently, here, contextual information is also considered and each class is not associated only to similar values in the color channels but also to a texture. Our approach is designed to ease the production of training samples for satellite image segmentation and the k-textures architecture could be adapted to support different number of bands and for more complex self-segmentation tasks, such as object self-segmentation. The model codes and weights are available at \url{https://doi.org/10.5281/zenodo.6359859}
\end{abstract}


\section{Introduction}

In recent years, automatic extraction of cartographic features from satellite images is undergoing a revolution thanks to advances brought by deep convolutional neural networks (CNNs). The main advantage of these supervised CNNs is that they take raw data and automatically learn features through training with minimal prior knowledge about the task \citep{lecun1998}. Furthermore, CNN accuracy in remote sensing applications is similar to human-level classification accuracy, but is consistent and fast, enabling rapid application over very large areas and/or through time \citep{BRODRICK2019734}. CNNs could support fast acquisition of accurate spatial information about land use and land cover change, which is essential for country planning and monitoring. However, among most significant efforts that are being made to map Land Use and Land Cover (LULC) such as the project MapBiomas, which maps LULC change specifically for Brazil \citep{Mapbio2018} or Global Forest Change product at 30 m spatial resolution of the University of Maryland \cite{Hansen850}, the CNNs are still under-used and semantic classifications rely mostly on machine learning algorithm that have been almost abandoned in the computer vision field -- such as the random forest algorithm -- because of lower accuracies when compared to CNNs and because the value of the reflectance is not so important for most application of deep learning in computer vision.

The use of supervised CNN techniques for these remote sensing applications is limited by their need of large training sample labelled by experts. Making manual samples, and even dense labelling (each pixel attributed to a label) is feasible in sub-metric spatial resolution images where objects such as buildings or trees are visible. However, this is more difficult and inaccurate for small objects at high or medium spatial resolution (5 to 30 m resolution) satellite imagery, such as Planet, Sentinel or Landsat, which are typically used for LULC change mapping at scale from regional to global. As consequence, manual sampling is generally made of spatial referenced points with labels determined either with high-resolution imagery or in the field. These sets of points are used to train machine learning (ML) algorithms, such as Random Forest, and to validate classification results. After the training step, in prediction, a class is attributed to each pixel of the satellite image scene. The classification process assumes that pixels are spatially independent. To improve the spatial coherence of the prediction, which are known to be noisy, additional temporal filters are used to improve the classification. For example, a pixel labelled as forest can't switch labels and then become forest again \cite{souza2020r}.

Several classical unsupervised algorithms are used to produce semantic segmentation in remote sensing classification of medium spatial resolution images such as k-means, ISOSeg, Maximum Likelihood, Mahalanobis Distance and Bhattacharyya Distance \citep{richards1999,camara2008}. However, they are not accurate enough to replaced manual labelling approach. This is easy to understand since they rely on clustering pixels with similar values in the colour channels, without considering spatial information of the neighbour pixels, which is over simplistic and naive.  For example, colors (or values in the colour channels red, green, and blue) can be seen as one feature when we currently know that several features and multiple levels of abstraction are need to reach state-of-the-art accuracy of classification \citep{lecun2015}, as demonstrated by the success of CNNs in computer vision tasks. Furthermore, while for natural object the color is an important feature, as it links structural and chemical composition \citep{Asner2015,Ferreira2016},  for most of non-natural objects the color is most of the time an undesirable feature. For example, to recognise cars or plastic balloons in images, color is not the most helpful feature, and often the CNN models are even trained with images where the hue is artificially changed (during data augmentation) to impede the model to give too much importance to the color. However, all these non-CNN classical unsupervised algorithm techniques are still used for classifying the high to medium spatial resolution satellite imagery because deep learning equivalent of these unsupervised models are still missing.

Unsupervised deep learning is seen as the future of deep learning, as this is mostly how human and animal learn, by observations  \citep{lecun2015}. This field of research has gained interest in the last years, and the term 'self supervision' is now preferred to 'unsupervised' because the model create its own abstraction of objects. Self-supervised image segmentation, in remote sensing or in computer vision, is currently one of the most challenging tasks for deep learning. Recent works show that self-supervised deep learning models can be trained in a way that semantic information emerges directly inside the features of the  model such as the self-supervised  Vision Transformer \citep{caron2021}. Other attempts to automatically segment images have used Variational Autoencoder (VAE) or simpler CNNs to produce the features and used k-means clustering on the features created by the algorithm to provide pseudo labels \citep{girard2019,Kim2020}. Finally, a last method is redrawing, where the model extracts an object's mask and redraw the object at the same location \citep{chen2019}. However, all of these unsupervised segmentation techniques, and more generally all the models of clustering with deep learning \citep{karim2021}, belong to soft clustering, that is, every pixel have a probability or likelihood to belong to each of the clusters. Clusters are obtained by the argmax function on the softmax layer of the object class, but are not used directly inside the model during optimization. 

By contrast, in hard clustering, the pixels belong or not to the cluster, 0 or 1. This problem of discretization using gradient descent is currently not known to be solvable by deep learning, because it is impossible currently to compute the gradient for discrete search. In other words, having a function that returns 0 or 1 (or more generally a step function) is discrete and non-differentiable and is not optimized by current gradient-based deep learning techniques. However, here, we present a simple technique to avoid this limitation, and to render a discrete search possible, that is, in our case, to assign a hard cluster to each class and use it in the gradient computation but without disrupting the gradient descent. There have been some works on the binarization of the weights in the deep learning field \cite{Qin2020} and some very close to the methods used here \cite{sakr2018,courbariaux2016}. However, the objective of theses works was to built lighter and more computationally efficient networks, not to use directly the binary weights as labels to provide the segmentation of an image.

Our model is called k-textures as it self-segments the image in $k$ hard clustering classes of different textures. As a study case, it was tested for self semantic segmentation of a cloud free RGB-NIR 20 km $\times$ 20 km Planet satellite image at 4.78 m of spatial resolution covering a degraded region of the Amazon forest in Brazil. The Planet images over the tropics image have been made available by the Norway’s International Climate and Forest Initiative (NICFI, \url{https://www.nicfi.no/}) \citep{Planet2017}. The results of k-textures were compared to the results obtained by k-means on the same data, which is the closest non deep learning machine learning algorithm for clustering image. Furthermore, we compare the obtained classes to the real world in land cover types obtained from MapBiomas \citep{Mapbio2018} to understand the capabilities and limitations of the k-textures model.

\section{Proposed K-textures Model}

\subsection{Hard sigmoid activation function with a very steep linear part}\label{sec_hs}

\begin{figure}[ht]
\begin{tikzpicture}
    \draw (0, 0) node[inner sep=0] {\centering\includegraphics[width=0.55\linewidth]{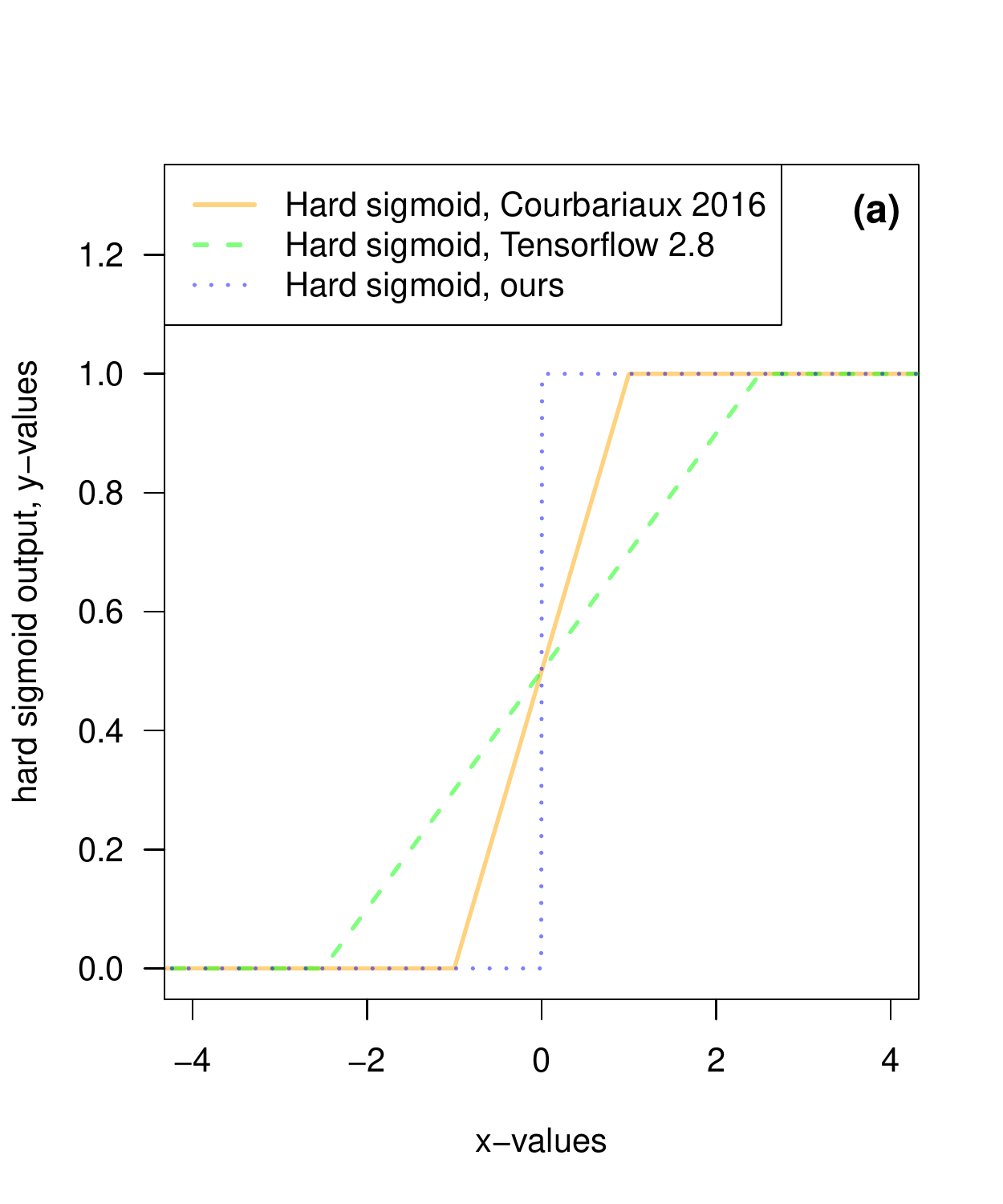}};
       \draw (0.475\linewidth, 0) node[inner sep=0] {\centering\includegraphics[width=0.35\linewidth]{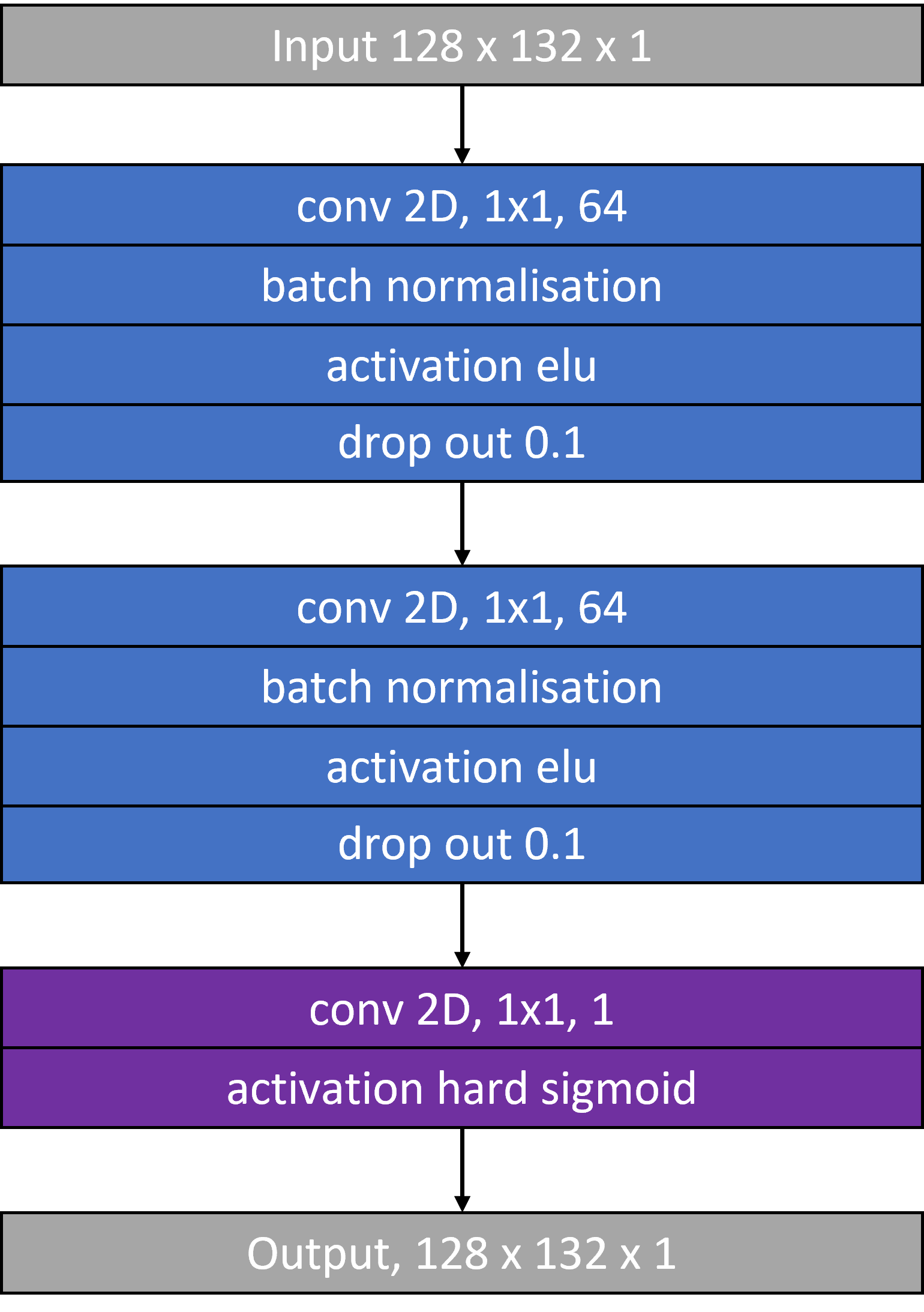}};
       \draw (0.68\linewidth, 0.23\linewidth) node {\large\textbf{(b)}};
\end{tikzpicture}
 \caption{Hard sigmoid curves for x values ranging between -4 and 4 for the function of \citet{courbariaux2016}, Tensorflow 2.8 \citep{AbadiAgarwalBarhamEtAl2015} and our function where the linear part rises from 0 to 1 for x values between 0 to 0.0002 in panel (a). Panel (b) : Architecture of the hard sigmoid approximation $\sigma_{CNN}$ model. Note that all the convolution layers have kernels of 1x1 indicating that each value is independent of the neighbors.}
 \label{Fig_1}
 \end{figure}

 A step function that returns only 0 or 1 from a continuous input variable is not differentiable and, thus, not usable in deep learning with gradient descent optimization. However, a differentiable function that returns 0 or 1 for most of the input values should exist. The hard sigmoid function is amongst the functions that are differentiable and used as activation function in deep learning and has this characteristic. The hard sigmoid function $\sigma(x)$ was originally defined as in eqn. \ref{eqn1-1} \citep{courbariaux2016} and Fig. \ref{Fig_1}. The function is composed of three connected parts, one that is constant and equal to 0, one that is linear and rises from zero to 1 and a final part, constant and equal to 1. For example, in Tensorflow (\url{https://www.tensorflow.org/api_docs/python/tf/keras/activations/hard_sigmoid})\citep{AbadiAgarwalBarhamEtAl2015} the hard sigmoid function \texttt{tf.keras.activations.hard\_sigmoid} return 0 if x $<$ -2.5, return 1, if x $>$ 2.5 and if -2.5 $<=$ x $<=$ 2.5: return 0.2 $\times$ x + 0.5,  eqn. \ref{eqn1-2} and Fig. \ref{Fig_1}. The coefficients of the hard sigmoid function can be adjusted so that the slope can occur in a very limited range of $x$ values. In other words, the coefficients can be set so that the function returns as $y$ constant values of 0 or 1 on most of the $x$ values range and only returns $y$ values on the range (0,1) on a very narrow range of $x$ values, with the advantage that the function remains differentiable. For example, in our hard sigmoid activation function, we use the coefficients of eqn. \ref{eqn1-3} which constraint the linear part that rise from 0 to 1 between $x$ values of 0 and 0.0002 and Fig. \ref{Fig_1}. 
\begin{eqnarray}
\sigma_{Courbariaux}\left(x\right) &=& \max\left(0,\min\left(1, x \times 0.5 + 0.5\right)\right) \label{eqn1-1} \\
\sigma_{Tensorflow}\left(x\right) &=& \max\left(0,\min\left(1, x \times 0.2 + 0.5\right)\right) \label{eqn1-2} \\
\sigma_{ours}\left(x\right) &=& \max\left(0,\min\left(1, x \times 5000 + 0 \right)\right) \label{eqn1-3} 
\end{eqnarray}

\subsection{A CNN to approximate the hard sigmoid function}

\begin{figure}[ht]
\centering\includegraphics[width=0.9\linewidth]{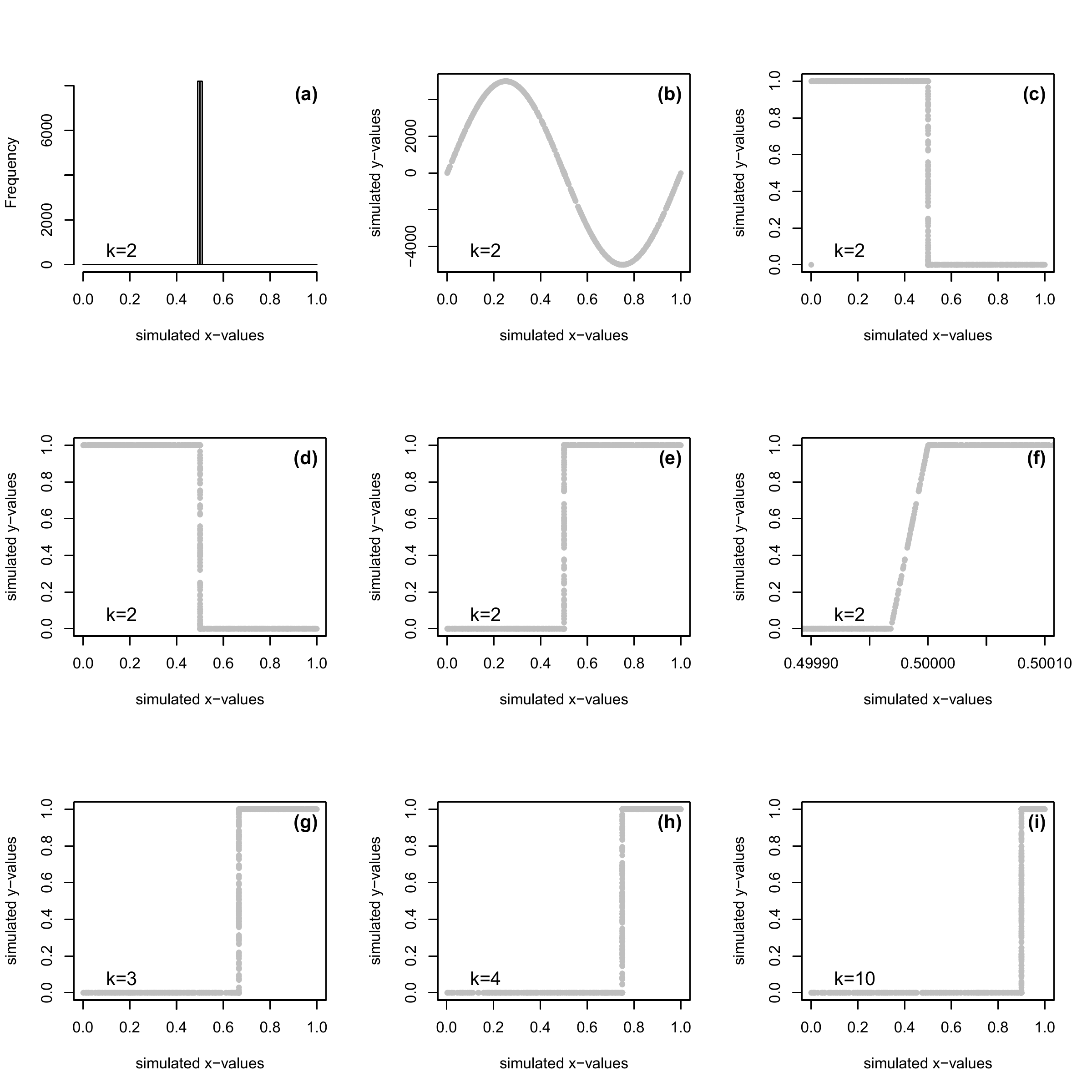}
 \caption{Data generated for training a CNN ($\sigma_{CNN}$) to approximate the hard sigmoid function with $k$ classes. The frequency of the simulated x values (eqn. \ref{eqnxsim1}-\ref{eqnxsim2}) for k=2 (a), simulated sin function from the simulated x values with eqn. \ref{eqnysim1} (b), clipped values on the range [0,1] eqn. \ref{eqnysim2} (c), y value different of 1 set to 1 for x value near 0 \ref{eqnysim3} (d), inversion along x for even $k$ classes \ref{eqnysim4} and y value different of 1 are set to 1 for x value near 1 \ref{eqnysim5} to obtain the final y value used to train the model for a number of classes $k$ = 2 (e), details of the linear part where y rise from 0 to 1 (f), examples of data simulated to train the model for a number of $k$ classes of 3 (g), 4 (h) and 10 (i).}
 \label{Fig_2}
 \end{figure}

As using the hard sigmoid function with such a steep slope directly inside a CNN would be instable, and, to ease the convergence, the hard sigmoid function was approximated by a CNN. This enables to decomposed the hard sigmoid function in a ensemble of small and easy to differentiate operations. To train a CNN that is able to approximate our hard sigmoid function, we build a small CNN architecture ($\sigma_{CNN}$) and simulated the data to train the model, Fig. \ref{Fig_1}b,  Fig. \ref{Fig_2} and eqn. \ref{eqnxsim0}-\ref{eqnysim5}. This CNN model is used to generate dummy variables for $k$ categories inside the k-textures model. Using a CNN allows us to decompose our problem in a series of small linear operations that can be  easily used as a layer with fixed weights inside a larger deep learning model, with the objective of transforming continuous data into categorical data while still enabling the use gradient descent optimization. The architecture of the hard sigmoid CNN model ($\sigma_{CNN}$) is presented in Fig. \ref{Fig_1}b. The model is made of two convolution blocks with 64 filters and 1 $\times$ 1 kernel, both consisting in the application of a 2D convolution layer, batch normalization, elu activation and dropout. Dropout was used to perform further implicit data augmentation and avoid overfitting during training. Finally, the output is produced with a final 2D convolution layer with 1 $\times$ 1 kernel, 1 filter and our custom hard sigmoid activation function (section \ref{sec_hs}).

The simulated input and output data to train the model were built with the functions described in eqn. \ref{eqnxsim0}-\ref{eqnysim5}, where $k$ is the number of classes, $k_{threshold}$ is the $x$ values located at the end of the rising linear part where $y$ reaches the value of 1, $x_{sim}$ are the simulated input values $y_{sim}$ are the simulated output values and $U$ a random uniform distribution for $n$ samples ranging from a defined minimum and maximum value. Input values were sampled from three uniform distributions with differences in frequency, that is, most of the input values (96.97\%) were sampled in an random uniform distribution located in the linear part of the curve, eqn. \ref{eqnxsim1}. For the two other distributions, the same number of values were simulated on each side of the linear part of the hard sigmoid curve to ease convergence, eqn. \ref{eqnxsim1}. All the simulated input values in the interval [-0.001,1.001] were clipped to [0,1] to match the range of values returned by a sigmoid activation function and also to have data for $x=0$ and $x=1$.
 
To simulate the $y$ value, the starting point is a sine function, eqn. \ref{eqnysim1}, with a large amplitude (5000) computed from the simulated $x$ values normalized between -1 and 1. The result is then clipped to the interval [0,1]. Then, test/corrections are made to ensure that $y$ values are equal to 0 or 1 on the  extremes, near $x=0$ and $x=1$. A sine function was used to ease the production of simulated output for different values of $k$ classes. As a convention, we keep the constant part equal to 1 on the right, so the function is inverted along $x$ for $k$ classes that are even. Simulated output data for $k$ = 2 are presented in Fig. \ref{Fig_2}e-f, for $k$ = 3 in Fig. \ref{Fig_2}g,for $k$ = 4 in Fig. \ref{Fig_2}e and for $k$ = 10 in Fig. \ref{Fig_2}i.

\begin{eqnarray}
 k_{threshold} &=& (k-1) / k  \label{eqnxsim0}\\
    x_{sim}&=& (U(min=-0.001, max=k_{threshold},n= 128 \times 2), \nonumber\\
           &  & U(min = k_{threshold} - 0.003 ,max = k_{threshold} + 0.003, n= 128 \times 128)), \nonumber\\
           &  & U(min=k_{threshold}, max=1.001 , n=128 \times 2) \label{eqnxsim1}\\
x_{sim}&=& clip(x_{sim},0,1) \label{eqnxsim2}\\
y_{sim}&=&\frac{5000}{k-1} \times \sin (((\frac{\pi/2}{1/k}) \times ((x_{sim} \times 2) - 1))) \label{eqnysim1} \\
y_{sim}&=& \begin{cases}
      0, & \text{if}\ y_{sim}<= 0 \\
      1, & \text{if}\ y_{sim}>= 1 \\
      y_{sim}, & \text{otherwise}\\
    \end{cases} \label{eqnysim2}\\
     y_{sim}&=& \begin{cases}
      1, & \text{if}\ x_{sim} < (1/k)/2 \\
      y_{sim}, & \text{otherwise}
    \end{cases} \label{eqnysim3} \\
 y_{sim}&=& \begin{cases}
      1 + (y_{sim} \times -1) & \text{, if}\ k \, \text{is even}\\
      y_{sim}, & \text{otherwise}
    \end{cases} \label{eqnysim4}\\  
 y_{sim}&=& \begin{cases}
      1, & \text{if}\ x_{sim} > 1 - ((1/k)/2) \\
      y_{sim}, & \text{otherwise}
    \end{cases} \label{eqnysim5} 
\end{eqnarray}

The hard sigmoid approximation $\sigma_{CNN}$ model for each $k$ classes was trained with 16384 images of 128 $\times$ 132 $\times$ 1 that represent 276824064 simulated $x$ values and their correspondent simulated $y$ value. All $x$ and $y$ were simulated again at each epoch. The model $\sigma_{CNN}$ was trained for 5000 epochs with a batch size of 1024. During network training, we used a standard stochastic gradient descent (SGD) optimization with a learning rate of 0.0001. The loss function was designed as a sum of two terms: binary cross-entropy and the Dice coefficient-related loss. If the linear part that rise from 0 to 1 of the curve produced by the CNN was in the interval $\pm$ 0.0001 around the value $1-1/k$, the desired accuracy was achieved and the training was stopped. Otherwise, the training continued for 5000 more epochs until the accuracy was attained (and repeated if necessary). The training of the model with $k$ = 16 and 5000 epochs took approximately 27 minutes using a Nvidia RTX2080 Graphics Processing Unit (GPU) with an 8 GB memory.

\begin{figure}[ht]
\centering\includegraphics[width=0.50\linewidth]{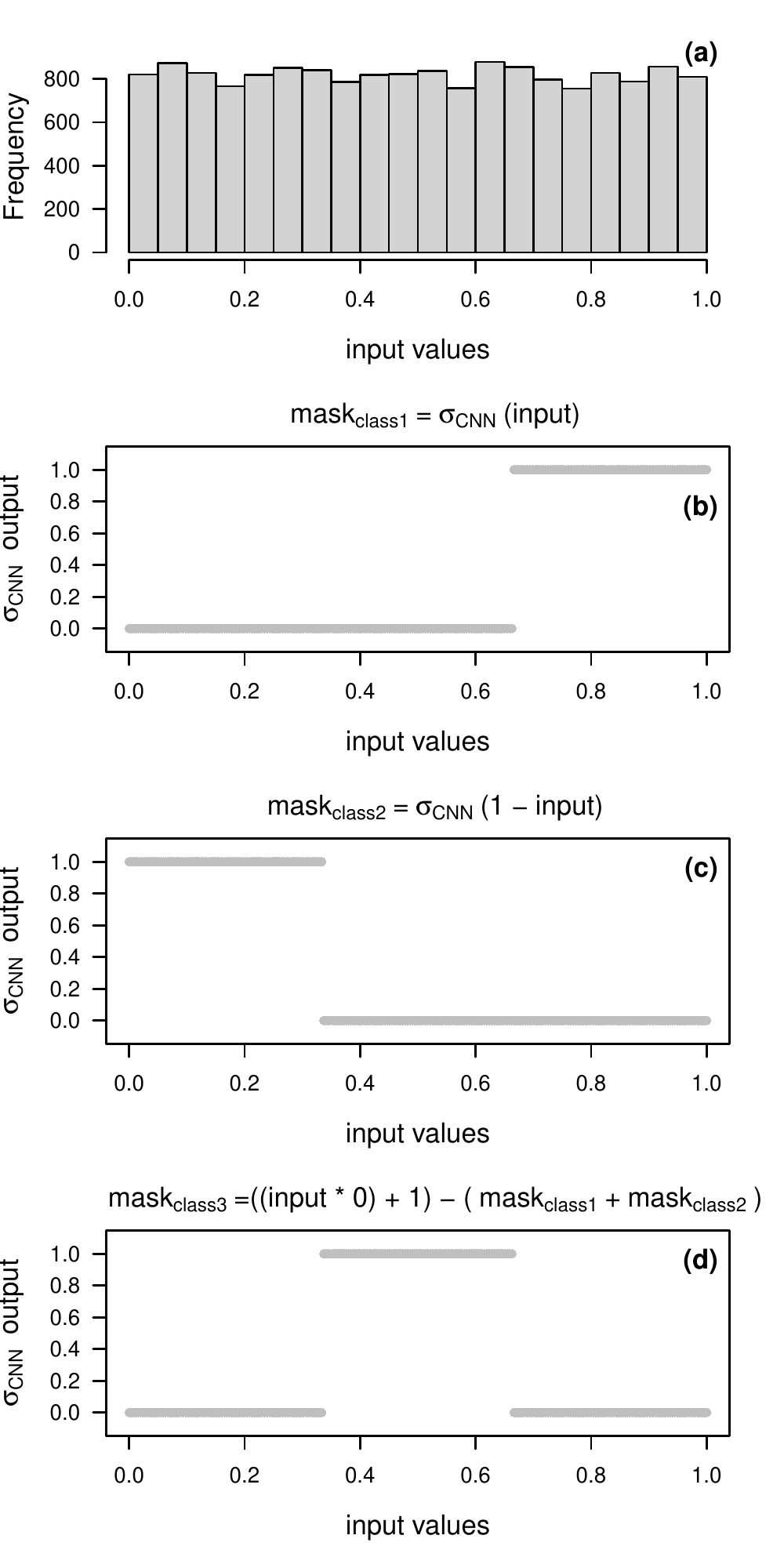}
 \caption{Computing binary mask values for $k$ = 3 with our $\sigma_{CNN}$ model which approximate a hard sigmoid function. Sample of uniform x values on the range [0,1] (a), values of the mask obtained by the binarization for mask$_{class1}$ (b),  mask$_{class2}$ (c) and mask$_{class3}$ (d).}
 \label{Fig_4}
 \end{figure}

The hard sigmoid approximation ($\sigma_{CNN}$) model with fixed weights corresponding to the $k$ class is used during the generation of the binary masks for each class in the k-texture model. For example, for $k = 3$ classes, the three binary masks can be computed with $\sigma_{CNN}$ for an input which values are in the range [0,1],  as presented in Fig. \ref{Fig_4}. The range [0,1] corresponds to the values returned by the convolution layer with a sigmoid activation function and here the example is given for input values on the range [0,1] with a uniform distribution, Fig. \ref{Fig_4}a. The first binary mask is computed by applying the $\sigma_{CNN}$ model directly on the input value, Fig \ref{Fig_4}b. The second binary mask is obtained by applying the $\sigma_{CNN}$ model to the (1 - input value), Fig \ref{Fig_4}c. The third binary mask is computed by removing the sum of the two previously obtained masks to a tensor the same size of the mask, filled of 1, Fig. \ref{Fig_4}d. The sum of the three masks is a tensor of ones. The operation to produce the binary masks can be generalized to any desired number of classes with a simple operation on the input values and the obtained intermediate masks (code available at \url{https://doi.org/10.5281/zenodo.6359859}).  Note that the values in the masks can be different of 0 or 1 if an x value is located on the linear segment that rises from 0 to 1 of the hard sigmoid produced by the $\sigma_{CNN}$ model. This issue is resolved later during the training of the k-texture model. Note that this method enables us to use the $\sigma_{CNN}$ model to transform a continuous into a binary variable and at the same time this model can be used with gradient descent because the $\sigma_{CNN}$ is fully differentiable.

\subsection{K-textures model architecture}

\begin{figure}[ht]
\centering\includegraphics[width=0.99\linewidth]{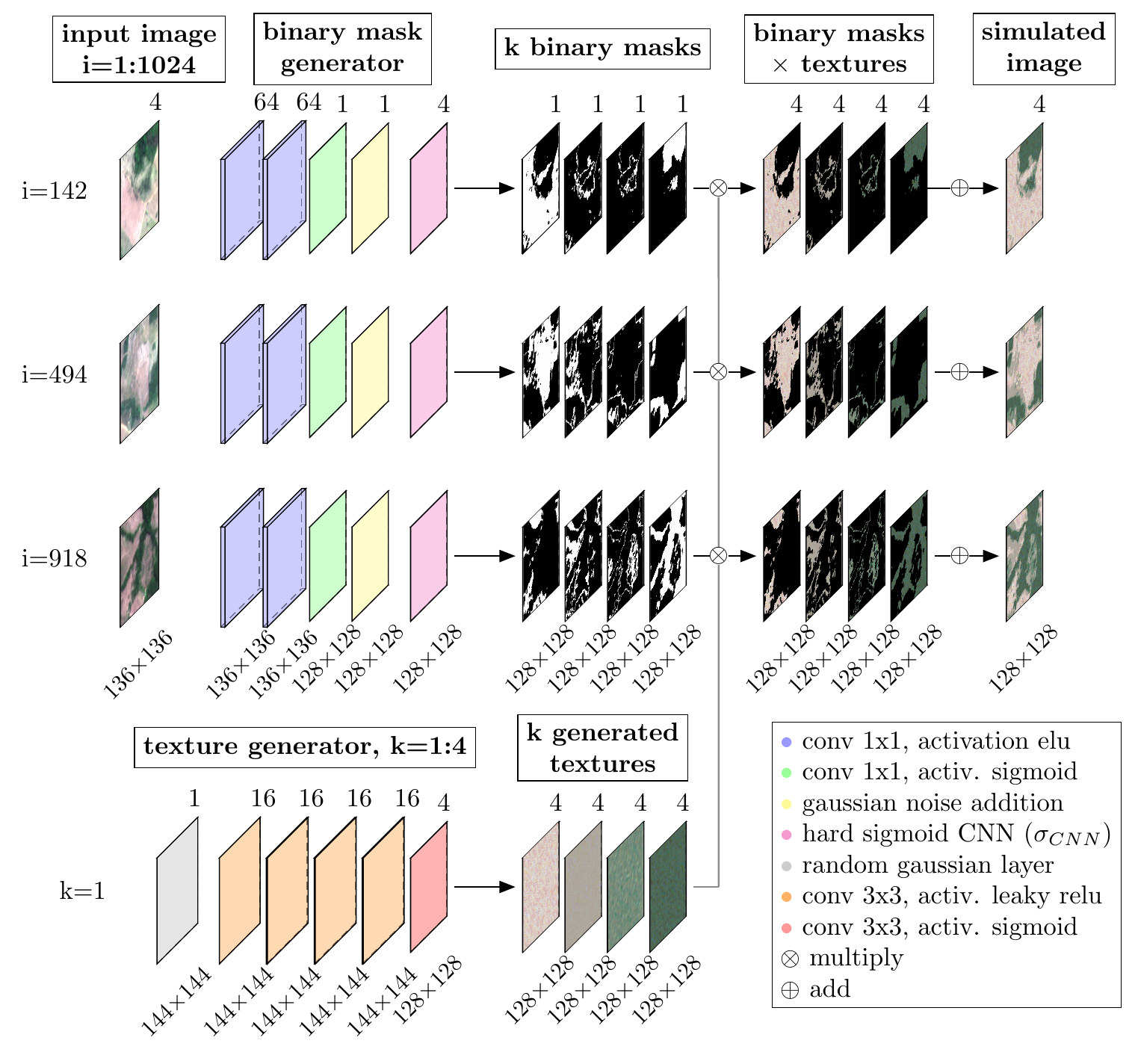}
 \caption{K-textures Model Architecture and example for $k$ = 4. The Planet images (4096 $\times$ 4096 pixels) is inputted to the k-textures algorithm in square patches of 136 $\times$ 136 pixels with 8 pixels of overlap (1024 patches). For each patches are generated $k$ binary masks. One texture is generated for each $k$ classes with the texture generator, to produce $k$ textures of 128 $\times$ 128 pixels. Note that only $k$ textures of 128 $\times$ 128 pixels are estimated for the entire Planet image, here $k = 4$ in our example. The $k$ textures and the $k$ binary masks are then multiplied and the result summed to produce a simulate image of each patch. }
 \label{Fig_5}
 \end{figure}

The objective of the k-textures algorithm is to self-segment an image in $k$ number of classes. The architecture of the k-textures algorithm is divided in two paths, Fig. \ref{Fig_5}, an encoder to generate the $k$ binary masks and an independent generator of textures. Both join before the end of the architecture to reconstruct the image.

The first path of the model is an encoder where the input image goes through two convolution blocks with 64 filters and 1 $\times$ 1 kernel (2D convolution layer, batch normalization and elu activation), and a last 2D convolutional layer which returns a tensor with sigmoid activation cropped to the size of (128,128,1), Fig. \ref{Fig_5}.
Kernels of  1  $\times$  1  are used in the encoder (each pixel is independent from the other), as CNNs using 3 $\times$ 3 kernel are already used in the feature extractor of the Loss function and there is no need to have them here (even if it is possible to use an encoder with 3x3 kernels). Furthermore, it enables to decrease the number of weights and it is more comparable to the k-means models where pixels are considered independently. To the sigmoid activation tensor returned by the encoder is added a Gaussian noise (sd = 0.0005), and the $\sigma_{CNN}$ model with fixed weights corresponding to the $k$ number of classes is applied to generate the $k$ binary masks.

The Gaussian noise amplitude (sd = 0.0005) is slightly larger than the $x$ value of the linear part of the $\sigma_{CNN}$ model that rises from 0 to 1. This adds instability to the gradient during training on the risen part of the hard sigmoid so that the algorithm tends to avoid these values. For example, on one epoch the x values  on the slope plus Gaussian noise can return a y value of 0 and 1 in the next epoch. With this simple Gaussian noise, the model is unable to learn the rising part of the curve and tend to avoid it because of its instability. Furthermore, and more importantly, the model is unable to use this part of the curve to propagate information, and, as a consequence, it tends to return either 0 or 1. On the other hand, the $x$ values located on the constant part of the curve (equal to 0 or 1) are stable even with the Gaussian noise.

The second path of the model is designed to generate $k$ textures, that is, one texture for each class. The texture is generated from a random Gaussian tensor of size (144,144,1). The texture is generated on a larger tensor than the original image to avoid the border effects. To the random Gaussian tensor is then applied a series of four identical convolutional blocks (2D convolution layer, batch normalization and leaky relu activation) with 16 filters and a kernel size of 3 $\times$ 3. Then a 2D convolution with 4 filters, a kernel size of 3 $\times$ 3 and a sigmoid activation is applied to produce the texture image. The generated textures are then clipped to the size of the binary masks.
Finally, the two paths meet to produce the reconstructed image from the $k$ binary masks and the $k$ textures. Each binary mask is multiplied by its correspondent textures, and all the resulting tensors are all summed together to produce the reproduced image, the output of the model.

The model is trained with the same image as input and as the image to reproduce, the only difference is that the input is slightly larger (136 $\times$ 136) than the image to reproduce (128 $\times$ 128) to avoid border effects. As these image are patches of a larger image, the larger border for the input is obtained by overlapping the neighbor patches and when overlapping is not possible, missing input data on the border is extrapolated by mirroring. The k-texture model is designed to optimize simultaneously using gradient descent, the $k$ textures and the $k$ binary masks that enable the production of an image the most similar to the original image. The model can be trained with various number of classes and we give the weights of the $\sigma_{CNN}$ model for the number of classes k from 2 to 64 but note that we only tested that the model works until k=32. We used the optimiser adam \citep{kingma2017} with a learning rate of 0.001 and the gradient norms were scaled if the gradient vector exceeds 1. The gradient scaling helps to stabilize the training of the model as was observed by trial and error. 

For the k-textures model, we designed a loss function which is the mean square error between the feature obtained from a feature extractor on the original image (input image) and the reconstructed image (model output). It enables the model to compare the two images not only on one feature, such as the color for example, but for all the different features present in the feature extractor. For the feature extractor, we use the pre-trained VGG16 model \citep{simonyan2014} with the fixed ImageNet weights available in Keras \citep{AllaireChollet,chollet2015keras}, specifically, the second to last layer of the model ($block5\_conv3$). The VGG16 feature extractor works only for RGB (3 bands) images, and, as we want to apply the model on 4 band images, we also train a custom variational auto-encoder (VAE) with 130 satellites images (Planet Scope, section \ref{satimg}) and a mean squared error (mse) loss to be further used as 4 bands image feature extractor.  The VAE model was a standard U-Net model \citep{Ronneberger2015} (\url{https://blogs.rstudio.com/ai/posts/2019-08-23-unet/}) with input of (256 $\times$ 256 $\times$ 4) but without skip connections in the decoder part. Note that other feature extractors could be used. The fixed weight used in the VAE for feature extraction were obtained after 3898 epochs when the VAE model had an accuracy of 0.9703441. The last central layer before the decoder of this VAE was used as feature extractor. The loss of the k-textures model is defined as the sum of the mean square error of the feature extracted with VGG16 and with our custom VAE.

All the models were coded in the programming language R \citep{CoreTeam2016} with Rstudio interface to Keras \citep{AllaireChollet,chollet2015keras} and Tensorflow \citep{AbadiAgarwalBarhamEtAl2015}.

\section{Experiments}

\subsection{Planet satellite image and associated land use/cover datasets}\label{satimg}

\begin{figure}[ht]
 \centering\includegraphics[width=1\linewidth]{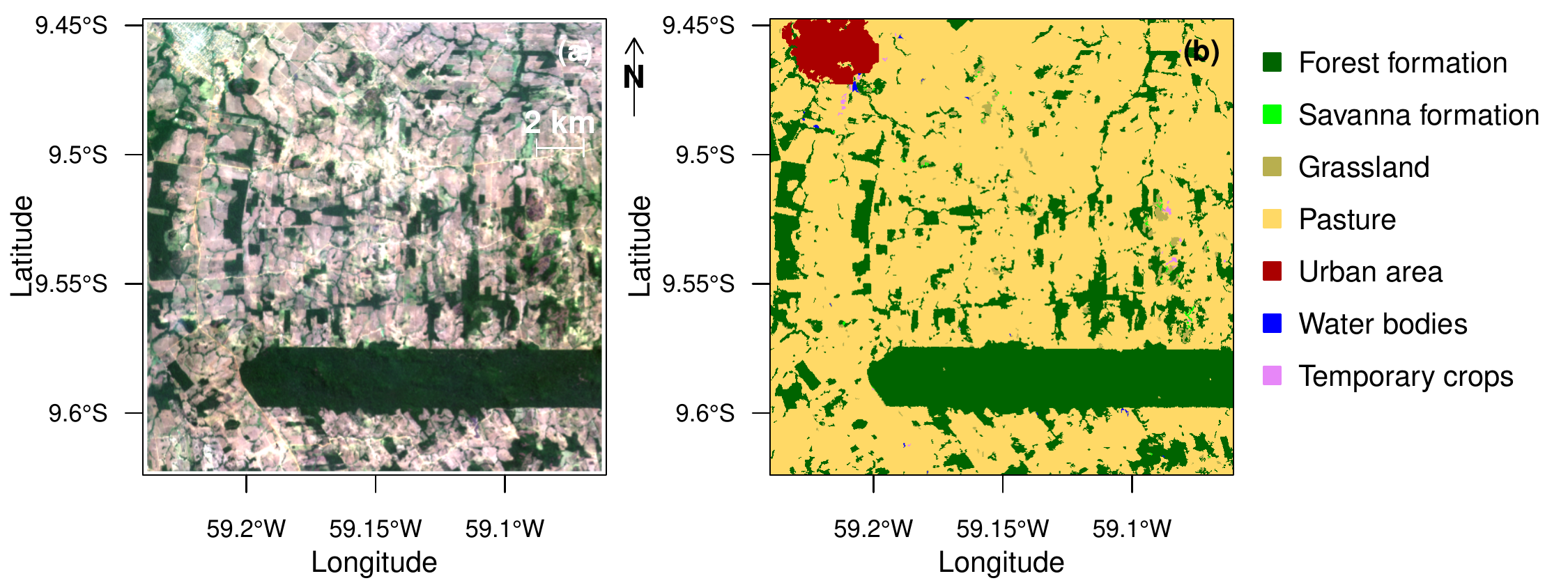}
  \caption{Geographical location of the region of interest with RGB composite of the Planet image used in the experiment (a); land use/cover classes of the region of interest for the year 2020 from the MapBiomas project (b).}
  \label{Fig1}
  \end{figure}

The experiment was undertaken in a region of the Brazilian Amazon forest located in the Mato Grosso State, Brazil, and centered at 9$\degre$32'8.43"S, 59$\degre$9'1.69"W, Fig \ref{Fig1}. The region contains several different land cover types, such as primary and secondary forests, pastures in use and abandoned and urban area. Furthermore this region is currently of critical importance and at the center of international attention because of the carbon emissions due to deforestation and forest degradation. It also contains numerous small patches of forests that would be highly time consuming to sample by hand in the high resolution image. Furthermore, in Brazil, the land cover types and water surfaces have been already mapped by the MapBiomas initiative using Landsat satellite image at 30 m spatial resolution, enabling us to understand and compare what the clusters of the k-textures and of the k-means models represent in the real world, Fig \ref{Fig1}b. 

The Planet image of 20 $\times$ 20 km at $\sim$ 4.78 m spatial resolution over the study region, Fig \ref{Fig1}a,  was downloaded trough the Planet API \url{https://api.planet.com/basemaps/v1/mosaics} \citep{Planet2017}. The image was selected because it has a lower cloud cover among the available images of the region times series and because it did not shown visible variation of atmospheric conditions or illumination effects.
The Planet ID of the image is ID\_ce7bad0f-a4a0-45fd-904b-eb6cc6eee373\_PAGE\_687-969\_DATE\_2021-08-01.tif and was acquired during July, 2021. We used 4 bands of this image: Red (0.650-0.682  $\mu$m), Green (0.547-0.585 $\mu$m), Blue (0.464 -0.517 $\mu$m) and the NIR bands (0.846-0.888 $\mu$m) \citep{planet2021}. All bands in raw image digital numbers (12 bits) were, first, truncated to the range 0--2540 for the RGB bands and scaled between 0 and 2540 for the NIR bands (i.e. divided by 3.937). Second, the 4 bands were scaled to 0--255 (8 bits) by dividing by 10 and then the Red-Green-Blue-NIR (RGBNIR) composite was built. No atmospheric correction was performed. A second image was generated from the composite adding a mirroring border of 4 pixels on each side for the generation of input images.

To test if our segmentation or the k-means based segmentation was consistent with independent datasets of land cover/use maps of the region, and if the classes seen by the k-texture model classes represent features in the real world, we compared the results to the land use/cover map from the Project MapBiomas - Collection 6 \citep{Mapbio2018}, Fig \ref{Fig1}b. MapBiomas Project is a multi-institutional initiative to generate annual land cover and use maps using automatic classification processes applied to satellite images. A complete description of the project can be found at \url{http://mapbiomas.org}. The MapBiomas land cover classes for the year 2020 over the region study extent at 30 m spatial resolution was obtained from google earth engine \url{https://mapbiomas.org/en/colecoes-mapbiomas-1?cama_set_language=en}. The MapBiomas image was resampled to overlay exactly the Planet image spatial resolution using \texttt{gdalwarp} with the nearest neighbour resampling method \citep{gdal2019}.

\subsection{Training}
 Clipping the Planet image in 136 $\times$ 136 pixels images with 4 pixels of overlap resulted in a sample set of 1024 images to train the model. The model output is a 128 $\times$ 128 image, however, using a 136 $\times$ 136 pixels image as input enabled to avoid border effects in the encoder output when using kernel of 3x3 pixels. This size of 136 $\times$ 136 pixels was selected because of the main texture of the land cover and land use classes is assumed to visible in image of that size (136 pixels $\sim$  649 m) and, because of GPU memory limitation, as a larger image would need larger GPU memory. Note that, for larger GPU memory, the algorithm could be adapted. We trained our network for 15000 epochs with a learning rate of 0.001, and where each epoch comprised 1 batch with 1024 images for the $k$ classes from 4 to 24 and 32. Training the model took $\sim$18 min for k=4 and $\sim$80 min for k=32 using GPU on an Nvidia RTX2080 with 8 GB of dedicated memory. 
 
 \section{Result}
 
\subsection{Model convergence}
 
    \begin{figure}[ht]
   \centering\includegraphics[width=0.5\linewidth]{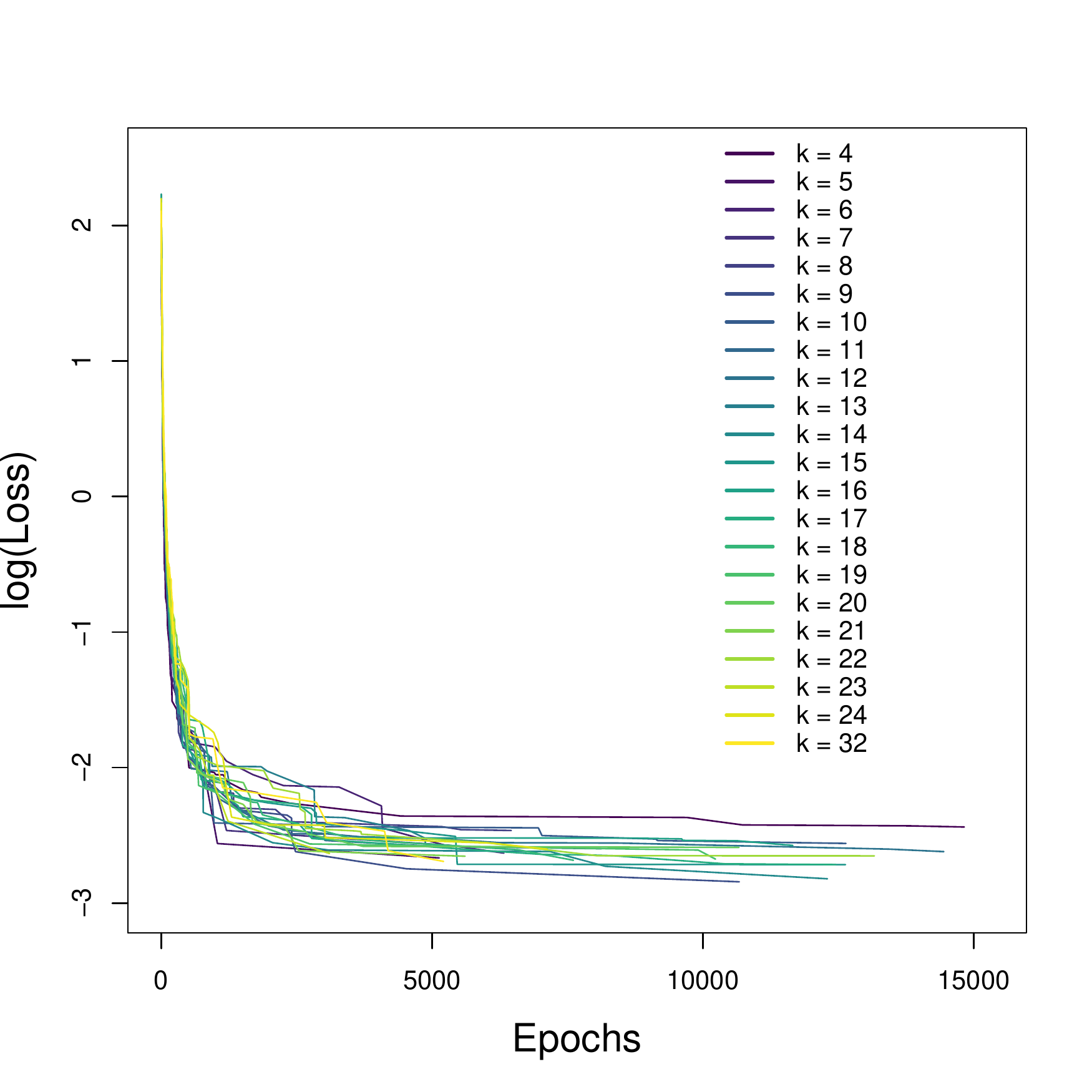}
 \caption{Evolution of the Feature Loss during the training of 15000 epochs for the $k$ considered classes, i.e. from 4 to 24 and 32. The loss line stops at the best loss value for the run with 15000 epochs and only the points with decrease in loss are represented. Loss is given in logarithmic scale to ease visualization.}
  \label{FigX}
  \end{figure} 
 
  The k-textures model converged as shown by the decrease of the loss during the 15000 epochs of training for all values of k, Fig. \ref{FigX}. The loss of the model for $k$ of 4 is the higher as expected and for other $k$ values lower loss values are observed but without clear order. The model shows sometimes relatively sharp changes in loss when improving, that could be due to the reorganisation of textures order or the creation of new textures/classes. The decrease in loss indicates that the model estimates the binary class and texture jointly to reproduce the original image.

\begin{table}[ht]
\centering
\begin{tabular}{r|rrr|rrr}
k & \multicolumn{3}{ |c| }{k-textures} & \multicolumn{3}{ |c }{k-means} \\ 
  \hline
 & actual\_k & loss & mae & actual\_k & loss & mae \\ 
  \hline
4 & 4 & 0.0873 & 0.0163 & 4 & 0.2328 & 0.0209 \\ 
  5 & 5 & 0.0695 & 0.0165 & 5 & 0.2027 & 0.0187 \\ 
  6 & 6 & 0.0757 & 0.0137 & 6 & 0.1851 & 0.0172 \\ 
  7 & 7 & 0.0720 & 0.0136 & 7 & 0.1375 & 0.0164 \\ 
  8 & 8 & 0.0850 & 0.0148 & 8 & 0.1267 & 0.0155 \\ 
  9 & 9 & 0.0582 & 0.0122 & 9 & 0.1192 & 0.0149 \\ 
  10 & 10 & 0.0773 & 0.0144 & 10 & 0.1138 & 0.0145 \\ 
  11 & 11 & 0.0744 & 0.0132 & 11 & 0.1048 & 0.0139 \\ 
  12 & 10 & 0.0728 & 0.0196 & 12 & 0.1011 & 0.0134 \\ 
  13 & 13 & 0.0777 & 0.0228 & 13 & 0.0957 & 0.0131 \\ 
  14 & 14 & 0.0596 & 0.0121 & 14 & 0.0890 & 0.0126 \\ 
  15 & 15 & 0.0661 & 0.0114 & 15 & 0.0863 & 0.0123 \\ 
  16 & 16 & 0.0762 & 0.0133 & 16 & 0.0845 & 0.0121 \\ 
  17 & 17 & 0.0661 & 0.0140 & --- &  ---  &  ---  \\ 
  18 & 17 & 0.0683 & 0.0135 & --- &  ---  &  ---  \\ 
  19 & 17 & 0.0689 & 0.0151 & --- &  ---  &  ---  \\ 
  20 & 20 & 0.0751 & 0.0202 & --- &  ---  &  ---  \\ 
  21 & 21 & 0.0704 & 0.0126 & --- &  ---  &  ---  \\ 
  22 & 20 & 0.0705 & 0.0135 & --- &  ---  &  ---  \\ 
  23 & 18 & 0.0718 & 0.0191 & --- &  ---  &  ---  \\ 
  24 & 23 & 0.0741 & 0.0158 & --- &  ---  &  ---  \\ 
  32 & 29 & 0.0677 & 0.0134 & --- &  ---  &  ---  \\ 
   \hline
\end{tabular}
\caption{Comparison between the k-textures and the k-means models of the number of clusters found (actual k), the feature loss and the mean average error. Feature Loss is the mean of individual losses. The results for k-textures were obtained on a run of 15360 epochs for each class.} 
\label{tbldp0}
\end{table}
 
 The k-textures model produces a better classification than the k-mean when looking at mean average error (mae) and Loss, Table \ref{tbldp0}. The k-textures model reaches lower mean average error (mae) with less k-classes when compared to k-means. For example, for 4 classes, the mae of k-textures is equivalent to the mae of 7 classes for k-means; and, for 9 classes, it already reaches the mae corresponding to 16 classes with k-means. This result was expected (and our experiments confirms it) since k-textures results involves several colors along with spatial information for one class while k-means only includes one color per class. For the loss, the reconstruction of the image with simulated textures also provides a resulting image that is closer to the original image, specifically, more similar in features than the image returned by the k-mean algorithm. This is also expected as is the value that the k-textures try to minimize. While k-means only focus on the values of pixels taken independently, the k-textures model uses CNNs and consequently multiple levels of abstraction and it can perform better.
  
While for the k-means, Loss and mae systematically improve when adding a class, the results of k-textures seem more subject to variations and to reach a limit where the model does not further improve. This effect could be related to the number of epochs, as 15360 epochs might not be enough for high number of groups, or other model parameters such as the batch size and learning rate. Further experiments will be needed to determine how to obtain the best model for a class. Sometimes, the k-textures model does not find the requested number of $k$ classes. In this case, the missing clusters are always those corresponding to the extreme values of the encoder sigmoid activation value (near 0 or near 1).  Finally, like the k-means, the k-textures results are always different between runs.

 \subsection{Binary masks}
 
\begin{table}[ht]
\centering
\begin{tabular}{rrrrrr}
  \hline
k & Number of 0 & Number of 1 & Number of non binary & Total number & percent non binary \\ 
  \hline
4 & 50331631 & 16777199 & 34 & 67108864 & 0.0000507 \\ 
  5 & 67108804 & 16777156 & 120 & 83886080 & 0.0001431 \\ 
  6 & 83886079 & 16777215 & 2 & 100663296 & 0.0000020 \\ 
  7 & 100663109 & 16777029 & 374 & 117440512 & 0.0003185 \\ 
  8 & 117440511 & 16777215 & 2 & 134217728 & 0.0000015 \\ 
  9 & 134217685 & 16777173 & 86 & 150994944 & 0.0000570 \\ 
  10 & 150994842 & 16777114 & 204 & 167772160 & 0.0001216 \\ 
  11 & 167772089 & 16777145 & 142 & 184549376 & 0.0000769 \\ 
  12 & 184549187 & 16777027 & 378 & 201326592 & 0.0001878 \\ 
  13 & 201326541 & 16777165 & 102 & 218103808 & 0.0000468 \\ 
  14 & 218103757 & 16777165 & 102 & 234881024 & 0.0000434 \\ 
  15 & 234880957 & 16777149 & 134 & 251658240 & 0.0000532 \\ 
  16 & 251657621 & 16776597 & 1238 & 268435456 & 0.0004612 \\ 
  17 & 268435275 & 16777035 & 362 & 285212672 & 0.0001269 \\ 
  18 & 285212603 & 16777147 & 138 & 301989888 & 0.0000457 \\ 
  19 & 301989392 & 16776720 & 992 & 318767104 & 0.0003112 \\ 
  20 & 318766787 & 16776899 & 634 & 335544320 & 0.0001889 \\ 
  21 & 335544143 & 16777039 & 354 & 352321536 & 0.0001005 \\ 
  22 & 352321512 & 16777192 & 48 & 369098752 & 0.0000130 \\ 
  23 & 369098492 & 16776956 & 520 & 385875968 & 0.0001348 \\ 
  24 & 385875506 & 16776754 & 924 & 402653184 & 0.0002295 \\ 
  32 & 520093579 & 16777099 & 234 & 536870912 & 0.0000436 \\ 
   \hline
\end{tabular}
\caption{Description of the values in the binary masks (0 , 1 or non binary values) which were obtained by the k-textures model for all the $k$ classes presented in this work (from 2 to 24 and 32).} 
\label{tblbin}
\end{table}

All the binary masks obtained after training for 15360 epochs are mostly filled with 0 and 1, that is, more than 99.999 \% of the values, Table \ref{tblbin}. This shows that our model produces stable binary masks that are effectively used as weights during the training with stochastic gradient descent. However, there were still a few non binary values and they represent a proportion ranging from 0.0000015 \% to 0.0004612 \% of the total pixels of the masks, Table \ref{tblbin}.  As the linear increasing part of the hard sigmoid returned by the $\sigma_{CNN}$ model ranges in a maximum interval of $\pm$0.0001 (0.0002) on the x-values. The percentage of expected values in this interval to be obtained by chance from a random distribution (ranging from 0 to 1 and with 4096$^2$ values like our tensor) will be of 0.02\%. Here, the observed percentage of non binary values is relatively smaller than this 0.02 percent, as the max percent of non binary values in our study is 0.0004612 \%. So the model is likely avoiding the linear increase. Actually, it cannot learn this part of the hard sigmoid, but still, due to the update of weights, some weights combination in the encoder can sometimes result by chance in a non binary value in the mask. So here, we acknowledge that we still cannot impede that a very limited number of values returned by the $\sigma_{CNN}$ model for the mask are non binary. In our case, for self segmentation, this is not so important, as even in the case of a non binary value for a pixel a unique cluster is attributed to the pixel with the argmax function. However, in further works which cannot have any non binary values, not even a very small percentage, this might be improved by increasing the slope of the hard sigmoid estimated by the $\sigma_{CNN}$ model.

\begin{figure}[ht]
\begin{tikzpicture}
    \draw (0, 0) node[inner sep=0] {\centering\includegraphics[width=0.20\linewidth]{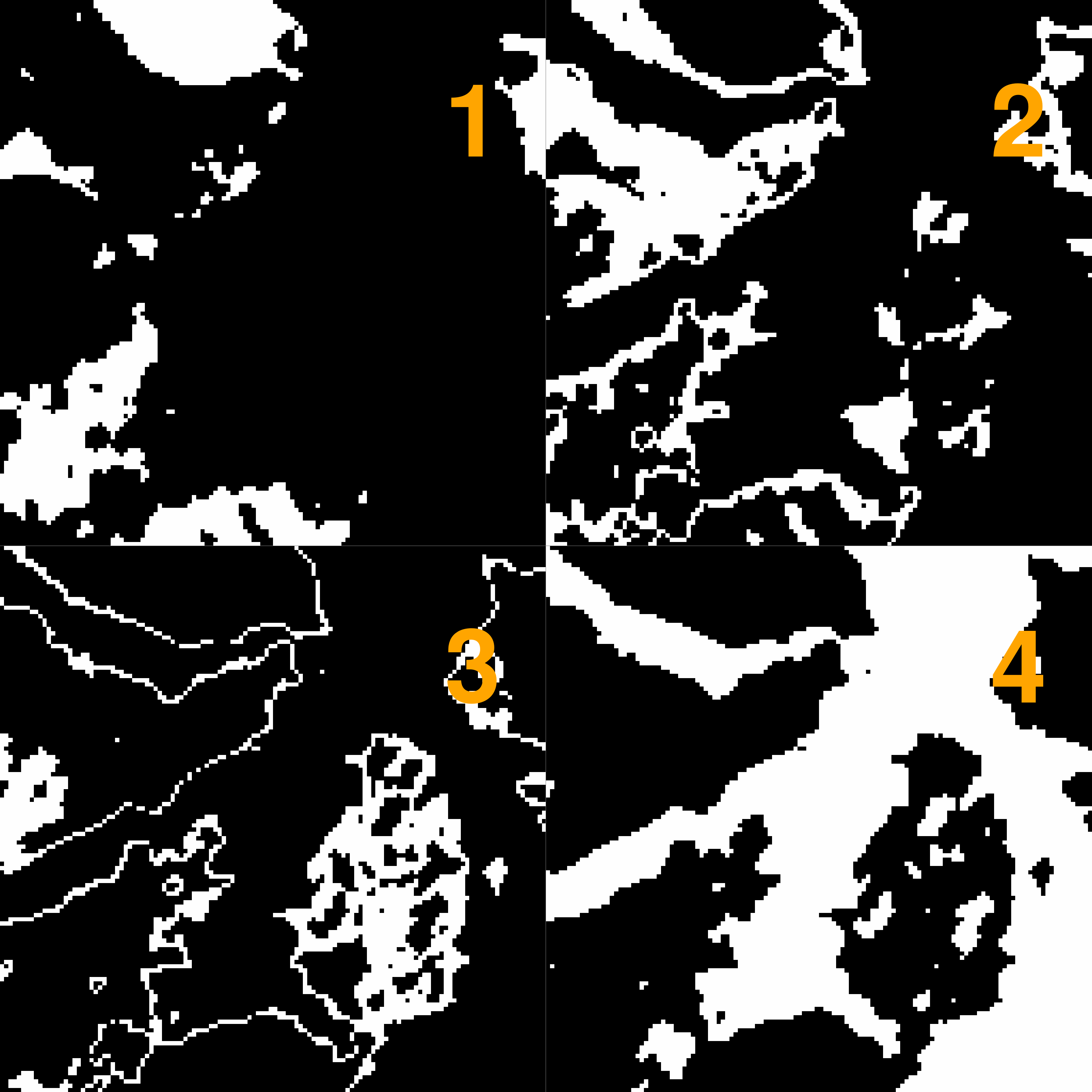}};
       \draw (0.20\linewidth + 0.10\linewidth, 0) node[inner sep=0] {\centering\includegraphics[width=0.30\linewidth]{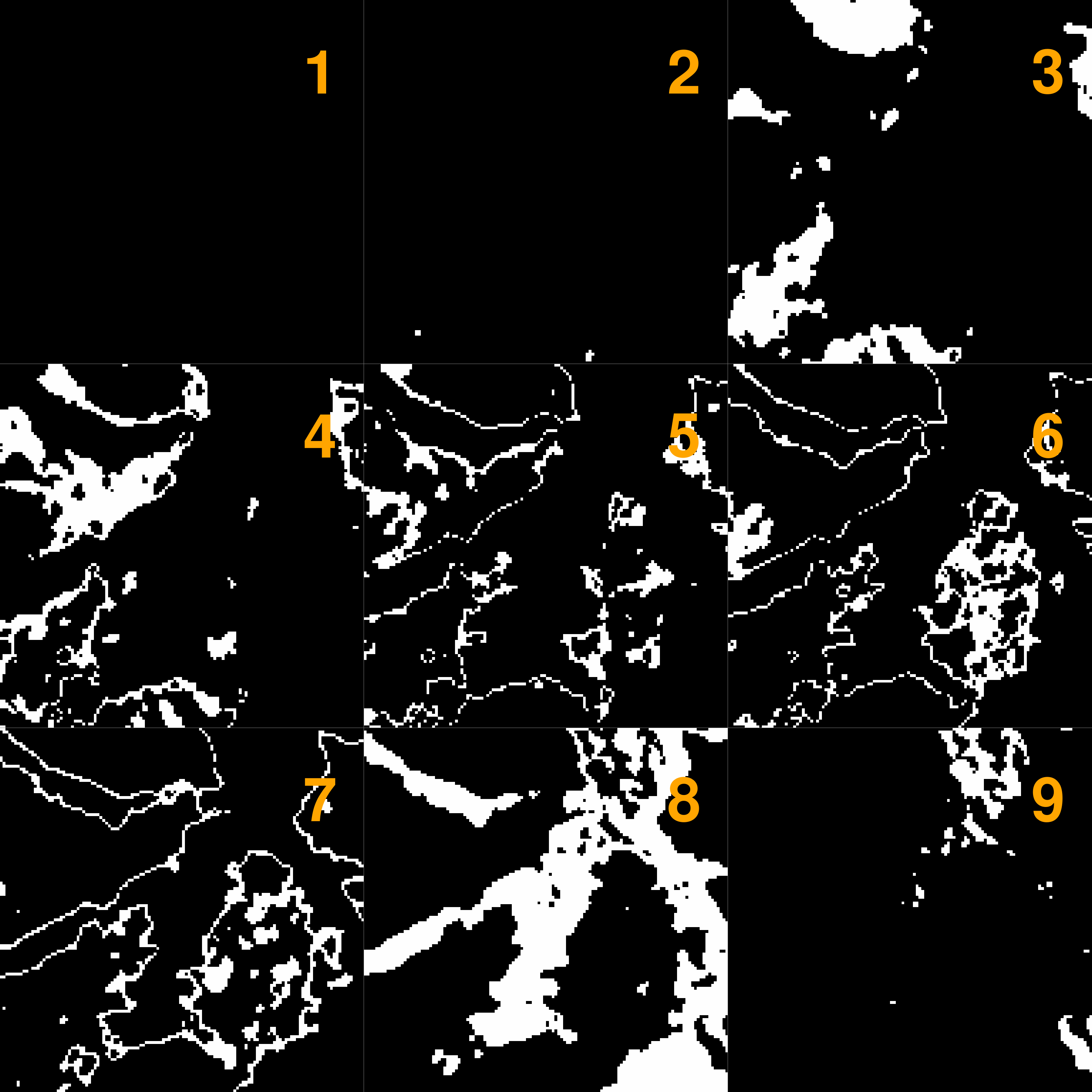}};
    \draw (0.40\linewidth + 0.20\linewidth +0.10\linewidth, 0) node[inner sep=0] {\centering\includegraphics[width=0.40\linewidth]{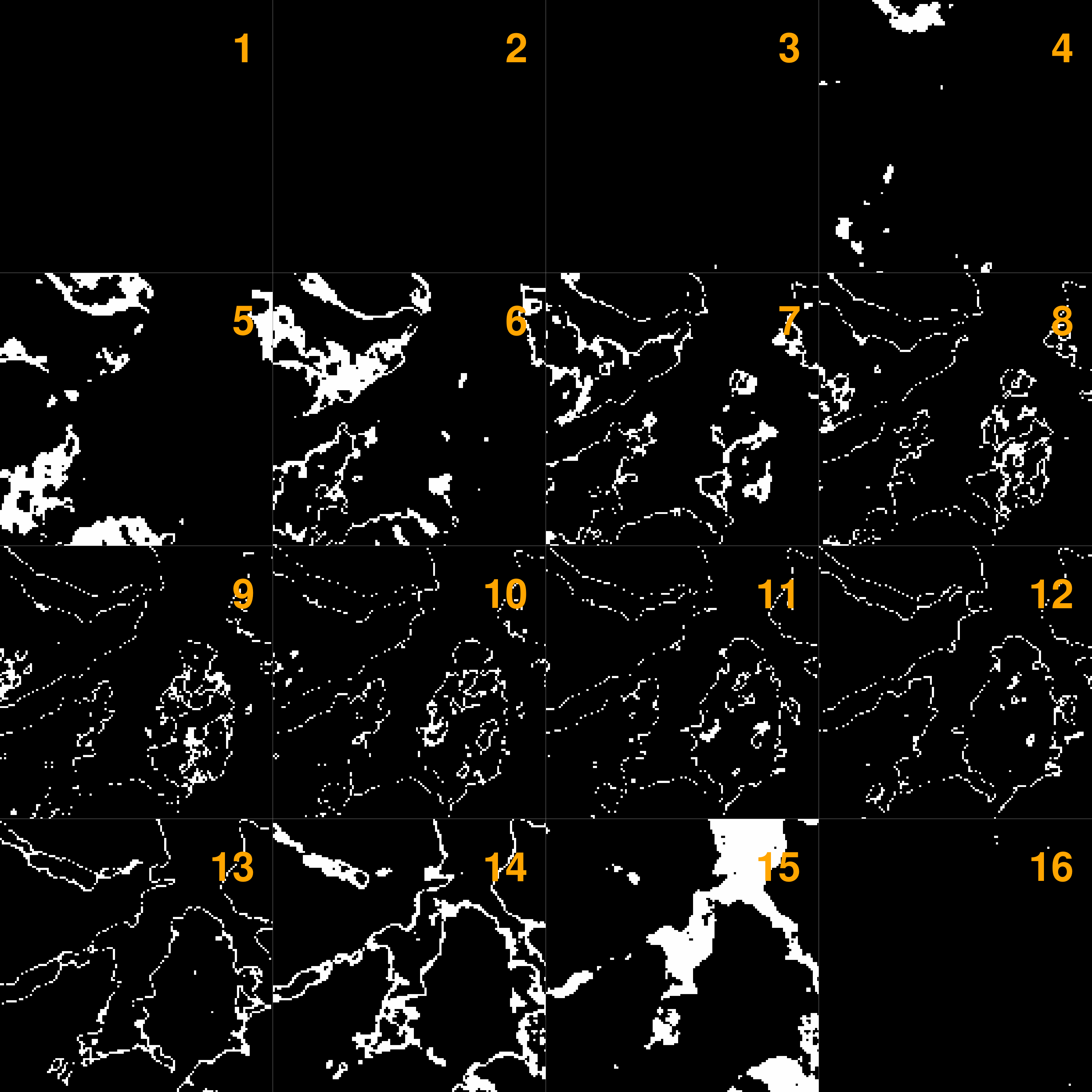}};
\draw (0, 0.10\linewidth + (0.025\linewidth) node {k = 4};
\draw (0.20\linewidth + 0.10\linewidth,  0.15\linewidth + 0.025\linewidth ) node {k = 9};
  \draw (0.40\linewidth +0.20\linewidth + 0.10\linewidth, 0.20\linewidth  + 0.025\linewidth) node {k = 16};
\end{tikzpicture}
  \caption{Example of binary masks obtained by the k-textures model for $k$ = 4, 9 and 16 classes for an patch of 128 $\times$ 128 pixels extracted from the planet 4096 $\times$ 4096 pixels image. For $k$ = 9 and $k$ = 16, the patch does not contain all the clusters,  it is not mandatory for a patch to contain all the clusters.}
  \label{Fig3}
\end{figure}

The binary masks provided by the model, for example in Fig. \ref{Fig3}, are estimated by the model and then use to multiply the corresponding textures that are also estimated in the same time. The final reconstituted image is then the sum of all the product of the binary mask and their correspondent texture, Fig. \ref{Fig_5}. A pixel of the image can have a the value 1 in only one binary mask. The values in the binary masks are weights of the models. Observing  Fig. \ref{Fig3}, it appears clearly that the binary mask can form coherent spatial pattern, that is the model can estimate the weights of the mask and of the texture in the same time, inside the same neural network. This shows that under certain condition, here in our case the restriction of the search space to a discrete search space, the CNN are able to converge even with some weights being only 0 and 1.

  \subsection{Generated textures}

  \begin{figure}[ht]
\begin{tikzpicture}
    \draw (0, 0) node[inner sep=0] {\centering\includegraphics[width=0.20\linewidth]{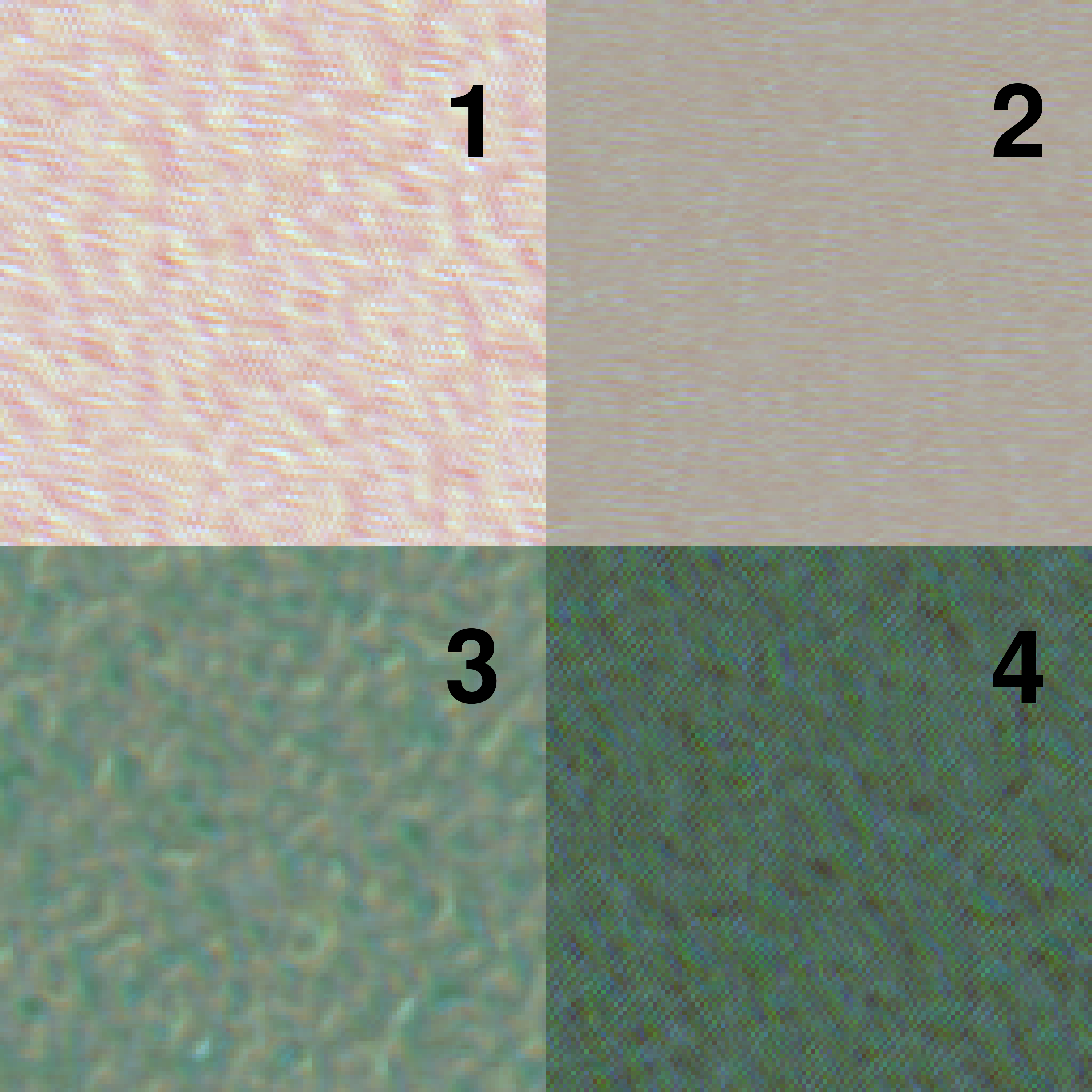}};
       \draw (0.20\linewidth + 0.10\linewidth, 0) node[inner sep=0] {\centering\includegraphics[width=0.30\linewidth]{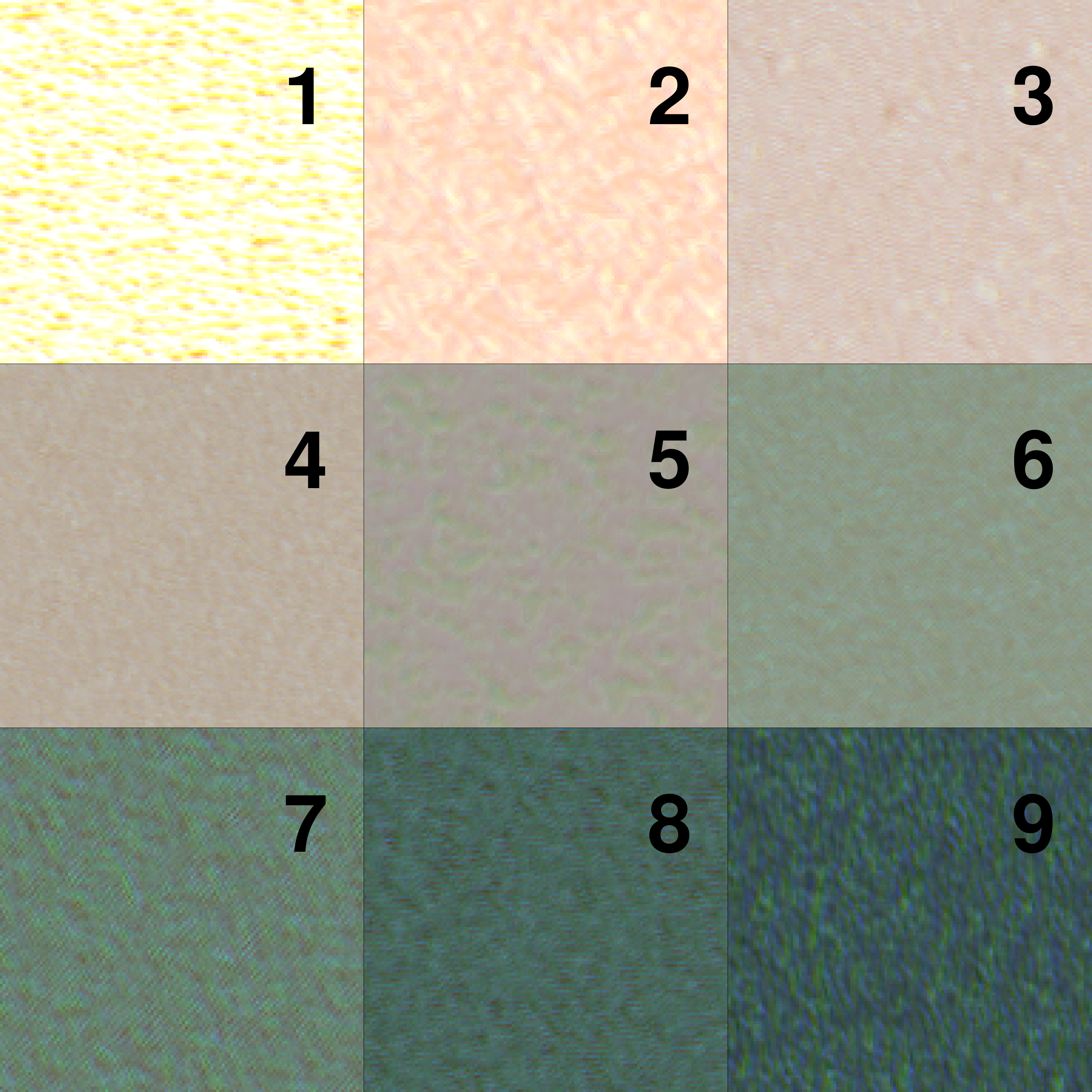}};
    \draw (0.40\linewidth + 0.20\linewidth +0.10\linewidth, 0) node[inner sep=0] {\centering\includegraphics[width=0.40\linewidth]{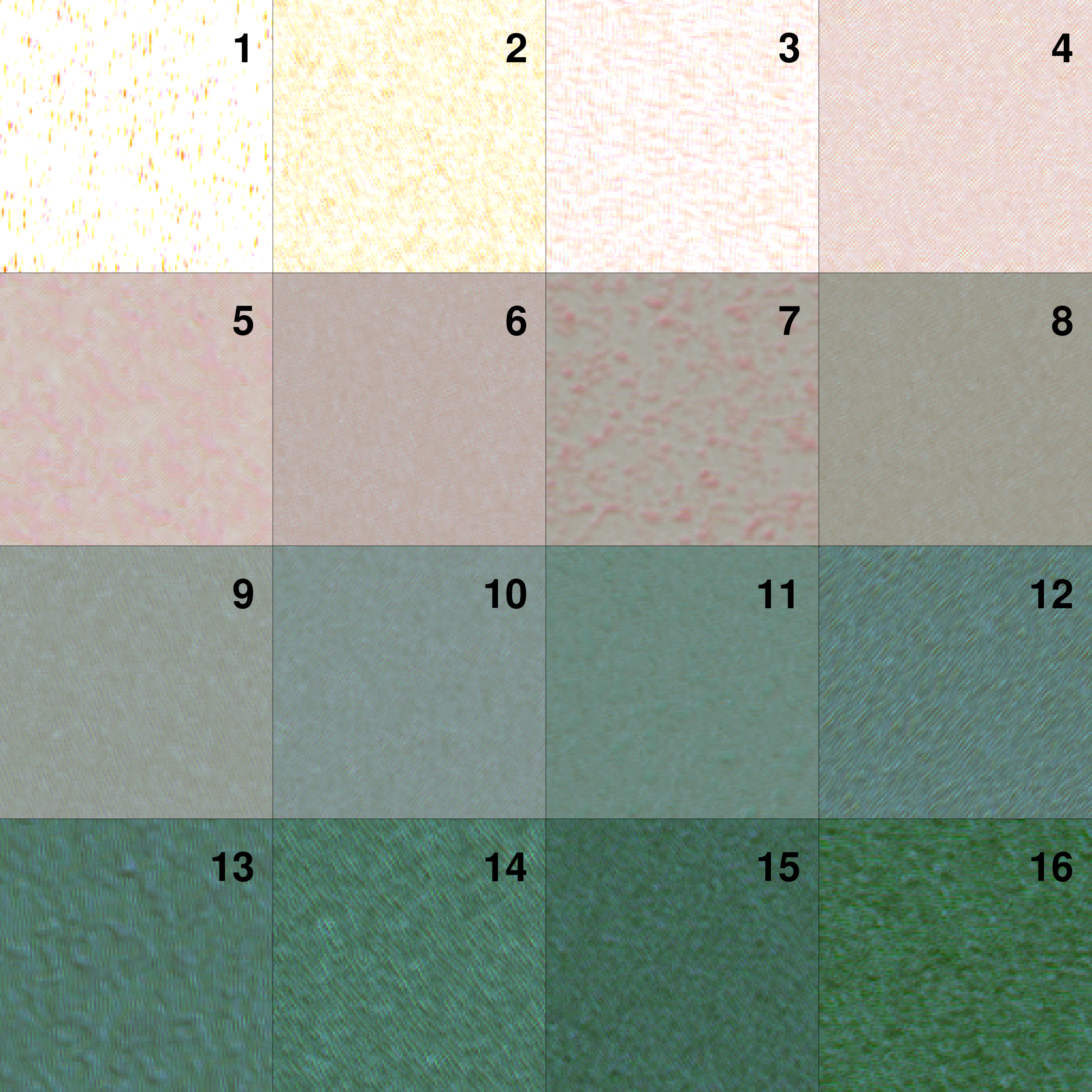}};
\draw (0, 0.10\linewidth + (0.025\linewidth) node {k = 4};
\draw (0, 0.18\linewidth + (0.025\linewidth) node {\huge \textbf{k-textures}};
\draw (0.20\linewidth + 0.10\linewidth,  0.15\linewidth + 0.025\linewidth ) node {k = 9};
    \draw (0.40\linewidth +0.20\linewidth + 0.10\linewidth, 0.20\linewidth  + 0.025\linewidth) node {k = 16};
\draw[] (-0.10\linewidth,0.20\linewidth  + 0.050\linewidth) -- (\linewidth-0.10\linewidth,0.20\linewidth  + 0.050\linewidth);
\draw[] (-0.10\linewidth, - 0.20\linewidth  - 0.050\linewidth) -- (\linewidth-0.10\linewidth, - 0.20\linewidth  - 0.050\linewidth);
\end{tikzpicture}
\begin{tikzpicture}
    \draw (0, 0) node[inner sep=0] {\centering\includegraphics[width=0.20\linewidth]{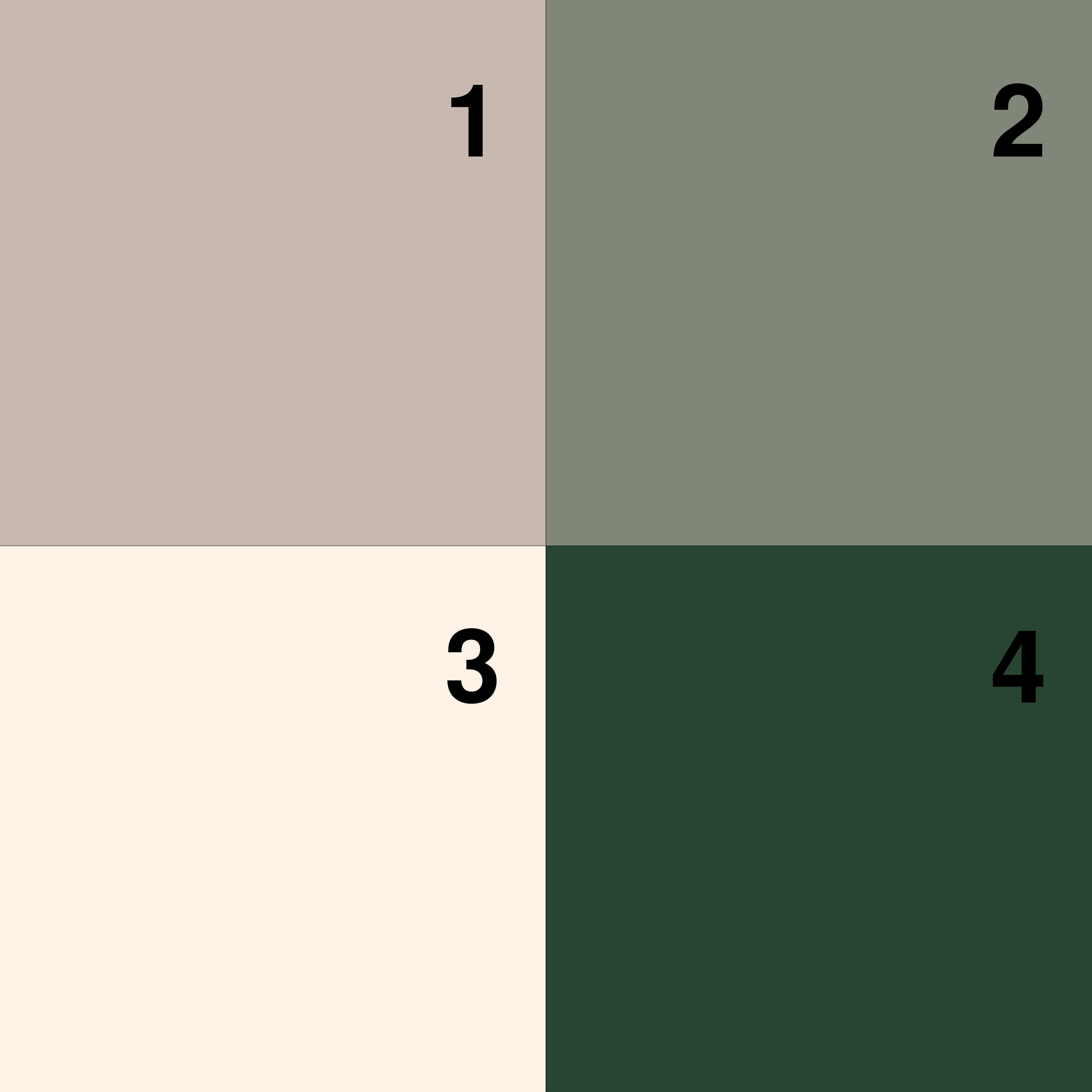}};
       \draw (0.20\linewidth + 0.10\linewidth, 0) node[inner sep=0] {\centering\includegraphics[width=0.30\linewidth]{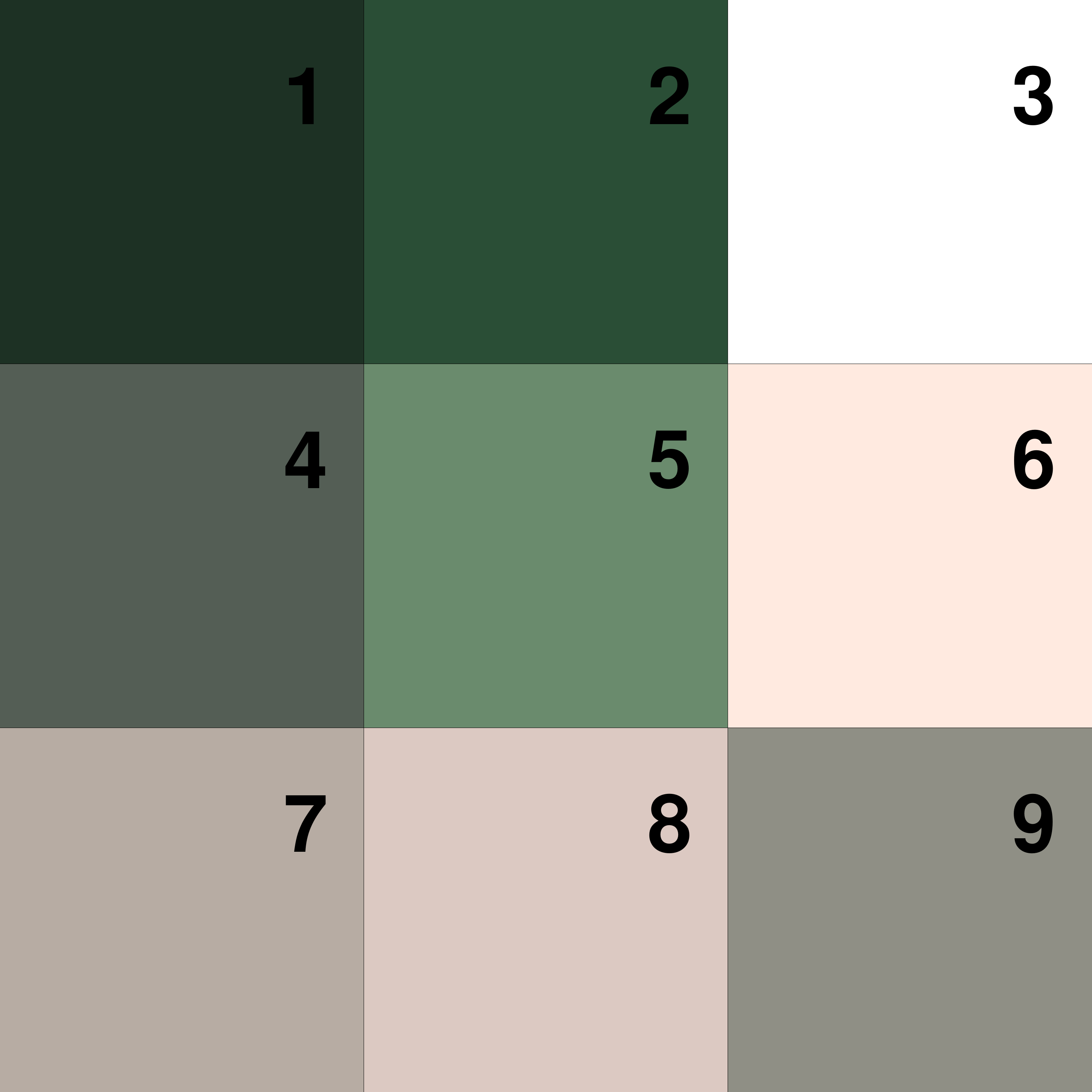}};
    \draw (0.40\linewidth + 0.20\linewidth +0.10\linewidth, 0) node[inner sep=0] {\centering\includegraphics[width=0.40\linewidth]{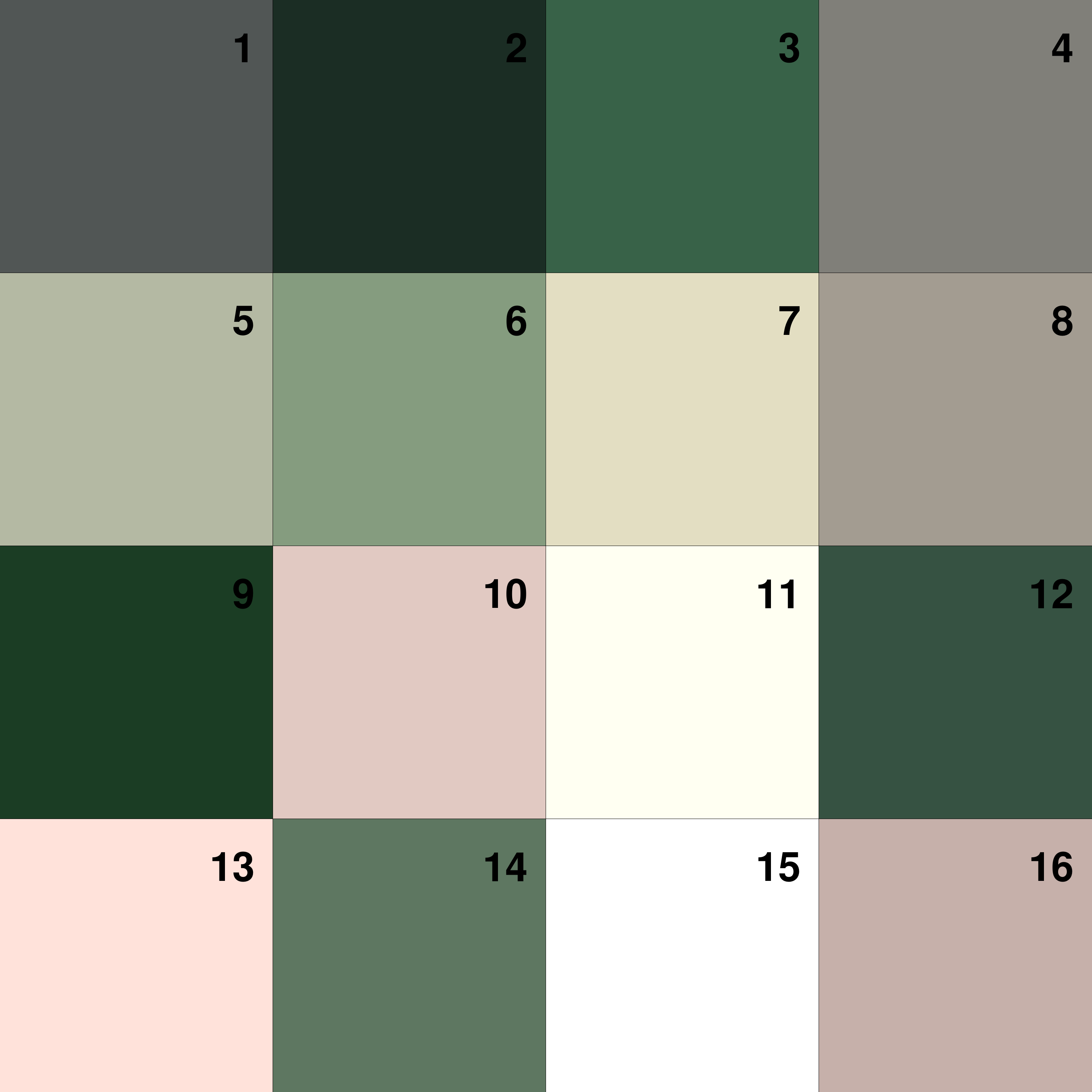}};
    \draw (0, 0.10\linewidth + (0.025\linewidth) node {k = 4};
    \draw (0, 0.18\linewidth + (0.025\linewidth) node {\huge \textbf{k-means}};
     \draw (0.20\linewidth + 0.10\linewidth,  0.15\linewidth + 0.025\linewidth ) node {k = 9};
    \draw (0.40\linewidth +0.20\linewidth + 0.10\linewidth, 0.20\linewidth  + 0.025\linewidth) node {k = 16};
 
\draw[] (-0.10\linewidth, - 0.20\linewidth  - 0.025\linewidth) -- (\linewidth-0.10\linewidth, - 0.20\linewidth  - 0.025\linewidth);
\end{tikzpicture}
  \caption{Example of the $k$ textures simulated by the k-textures model and the colors of the cluster obtained by the k-means for $k$ = 4, 9 and 16 classes. The matrix of textures are filled by row, starting by the top left, in the order returned by the models. Note that, for an easy visualization, only RGB bands are displayed but the estimated textures have 4 bands, RGB and NIR.}
 \label{Fig2}

\end{figure}

The $k$ textures simulated by the k-textures model and the $k$ colors of the cluster obtained by the k-means for $k$ in 4, 9 and 16 are given in Fig. \ref{Fig2}. The textures estimated by the $k$ textures model look more natural and organic than the monochromic image returned by k-means. The textures estimated for forest, for example, are visually close to what is observed in the real images, with brighter crowns and a lot of shade. The grain is different between the obtained textures showing that our texture generator is able to simulate a variety of textures. Some textures appear with light strides but it is still not clear if this is due to the generator, like it could be with GANs or because this is actually better for the Loss. 

Surprisingly, the k-textures model sorts the textures, with apparently less difference between the nearby textures, and with the textures of higher hue at the opposite of the textures lower hue. This 1D order likely emerged from the model architecture and was unexpected.  This could be due to the methods for producing the binary masks with $\sigma_{CNN}$, as the attribution to a class is only determined by the value ranging between 0 and 1 returned by the encoder. It seems that the model can improve the loss by ordering the textures, as this order is found for all values of k. Here the colors in the satellite image does not show extreme variations and are relatively close ($\sim$ terrains colors), but it could be interesting to see how the model behaves with more complex colors composition. Ordering color is a challenge in machine learning and here our algorithm manage to order the classes by textures in 1D and clustering at the same time. As a comparison, one of the best recent machine learning algorithms to represent color or classes on a 2D projection, t-SNe \citep{van2008}, is designed to order the pixels in a 2D plan to provide a representation of the data, but not to cluster the data.

For k-means, the obtained colours are similar to those in the textures produced by the k-textures, Fig. \ref{Fig2}. No order appears as the k-means is not made for this and only searches for centroids (in 4 dimensions in our specific case). Colors with higher and lower hues are observed in the classes returned by the k-means, while, for the k-textures model, the colors with highest or lowest hue appeared mixed in textures with other colors.

 \subsection{Comparison of k-textures classes with real world land use classes from Mapbiomas}
 
 The k-textures shows association with the real world land cover class from Mapbiomas, that is, most of the land cover classes are represented by a limited number of k-textures clusters, Table \ref{ktkmk9}. As our model orders the textures, it renders the table easy to read and to determine which cluster correspond to which class or to which transition between land cover classes, Table \ref{ktkmk9}. For example, from the results of a k-textures model trained using 9 classes, Table \ref{ktkmk9}, it can be observed that the forests are mostly in the cluster 8 and 9 of the k-textures model. Then the savanna, grassland and water bodies are in cluster 8, 7, and 6 and clusters lower than 5 are related to pasture and urban areas. For the temporary crops, some were in cluster of vegetation  (around cluster 7) and some in cluster without vegetation or cultures and with bare soils (cluster 2). The land cover class sometimes overlaps with two or three clusters, however note that the land cover classes are made with Landsat images at 30 m spatial resolution while the Planet images used in the k-textures models have a spatial resolution of $\sim$4.77 m, so some overlap is expected. The table of the k-means results does not bring much visual information, Table \ref{ktkmk9} and seems to associate one particular color to each class. In our results, Table \ref{ktkmk9}, the k-textures clusters are more unbalanced (0.07\% - 27.58\%) that the k-means clusters (2.55\% - 18.49\%) but this could be an artifact as k-means does not search for the same number of observations inside each cluster, and more investigation should be made to check if our model is able to cluster data situation not know to be easily resolved by k-means, such as when clusters are of different sizes and densities. %

\begin{table}[ht]
\tiny
\centering
\begin{tabular}{rrrrrrrrrrl}
  \hline
algorithm & cluster & Forest & Savanna & Grassland & Pasture & Urban   & Water & Temporary& Total  \\ 
 &  &  Formations & Formation &  &  & Area  & bodies & Crops &  cluster\\ 
 &  &  number (\%) & number (\%)  & number (\%) & number (\%) & number (\%)   & number (\%)  & number (\%)  &  number (\%) \\ 
  \hline
k-textures & 1 & 64 (0) & 0 (0) & 36 (0.02) & 7960 (0.07) & 3213 (0.89) & 0 (0) & 579 (3.82) & 11852 (0.07) \\ 
  k-textures & 2 & 4619 (0.11) & 2 (0.01) & 2240 (1.54) & 2498506 (20.69) & 135449 (37.31) & 947 (10.09) & 5037 (33.22) & 2646800 (15.78) \\ 
  k-textures & 3 & 23190 (0.56) & 78 (0.47) & 8522 (5.87) & 4481821 (37.12) & 111581 (30.74) & 861 (9.17) & 1459 (9.62) & 4627512 (27.58) \\ 
  k-textures & 4 & 27314 (0.66) & 277 (1.67) & 8590 (5.91) & 1580666 (13.09) & 39856 (10.98) & 603 (6.43) & 679 (4.48) & 1657985 (9.88) \\ 
  k-textures & 5 & 32008 (0.77) & 1225 (7.38) & 10631 (7.32) & 955682 (7.91) & 22656 (6.24) & 650 (6.93) & 776 (5.12) & 1023628 (6.1) \\ 
  k-textures & 6 & 51570 (1.24) & 4038 (24.33) & 20606 (14.19) & 849797 (7.04) & 18829 (5.19) & 1428 (15.22) & 1562 (10.3) & 947830 (5.65) \\ 
  k-textures & 7 & 134571 (3.24) & 6234 (37.57) & 38014 (26.18) & 898254 (7.44) & 17592 (4.85) & 1974 (21.03) & 2919 (19.25) & 1099558 (6.55) \\ 
  k-textures & 8 & 1818045 (43.78) & 4671 (28.15) & 49543 (34.11) & 780256 (6.46) & 13784 (3.8) & 2245 (23.92) & 2153 (14.2) & 2670697 (15.92) \\ 
  k-textures & 9 & 2061754 (49.64) & 70 (0.42) & 7046 (4.85) & 21763 (0.18) & 44 (0.01) & 677 (7.21) & 0 (0) & 2091354 (12.47) \\
   \hline
  k-means & 1 & 2147535 (51.71) & 602 (3.63) & 11069 (7.62) & 100767 (0.83) & 834 (0.23) & 559 (5.96) & 5 (0.03) & 2261371 (13.48) \\ 
  k-means & 2 & 1522473 (36.66) & 240 (1.45) & 1076 (0.74) & 357783 (2.96) & 8526 (2.35) & 19 (0.2) & 297 (1.96) & 1890414 (11.27) \\ 
  k-means & 3 & 1201 (0.03) & 0 (0) & 601 (0.41) & 364502 (3.02) & 56652 (15.61) & 515 (5.49) & 3520 (23.21) & 426991 (2.55) \\ 
  k-means & 4 & 294582 (7.09) & 10166 (61.26) & 81550 (56.15) & 917499 (7.6) & 11513 (3.17) & 3806 (40.55) & 4782 (31.54) & 1323898 (7.89) \\ 
  k-means & 5 & 84152 (2.03) & 184 (1.11) & 2120 (1.46) & 849102 (7.03) & 22710 (6.26) & 259 (2.76) & 524 (3.46) & 959051 (5.72) \\ 
  k-means & 6 & 2423 (0.06) & 1 (0.01) & 1068 (0.74) & 1641630 (13.6) & 63548 (17.51) & 316 (3.37) & 1833 (12.09) & 1710819 (10.2) \\ 
  k-means & 7 & 28413 (0.68) & 145 (0.87) & 9122 (6.28) & 2984896 (24.72) & 77664 (21.39) & 718 (7.65) & 905 (5.97) & 3101863 (18.49) \\ 
  k-means & 8 & 8468 (0.2) & 27 (0.16) & 4171 (2.87) & 2900254 (24.02) & 75749 (20.87) & 595 (6.34) & 1123 (7.41) & 2990387 (17.82) \\ 
  k-means & 9 & 63888 (1.54) & 5230 (31.52) & 34451 (23.72) & 1958272 (16.22) & 45808 (12.62) & 2598 (27.68) & 2175 (14.34) & 2112422 (12.59) \\ 
   \hline
\end{tabular}
\caption{Distribution of the clusters of k-textures and k-means by Mapbiomas landcover classes.} 
\label{ktkmk9}
\end{table}

  \subsection{Example of tropical forest self-segmentation}
\begin{figure}[ht]
  \centering\includegraphics[width=0.50\linewidth]{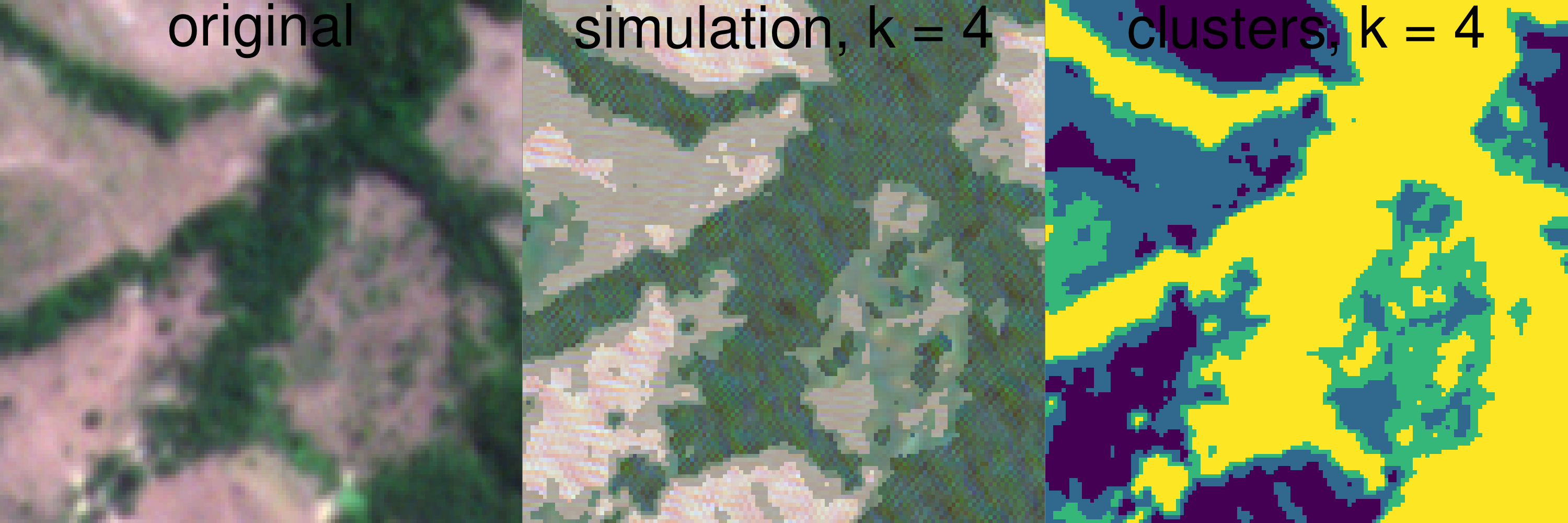}\\
  \centering\includegraphics[width=0.50\linewidth]{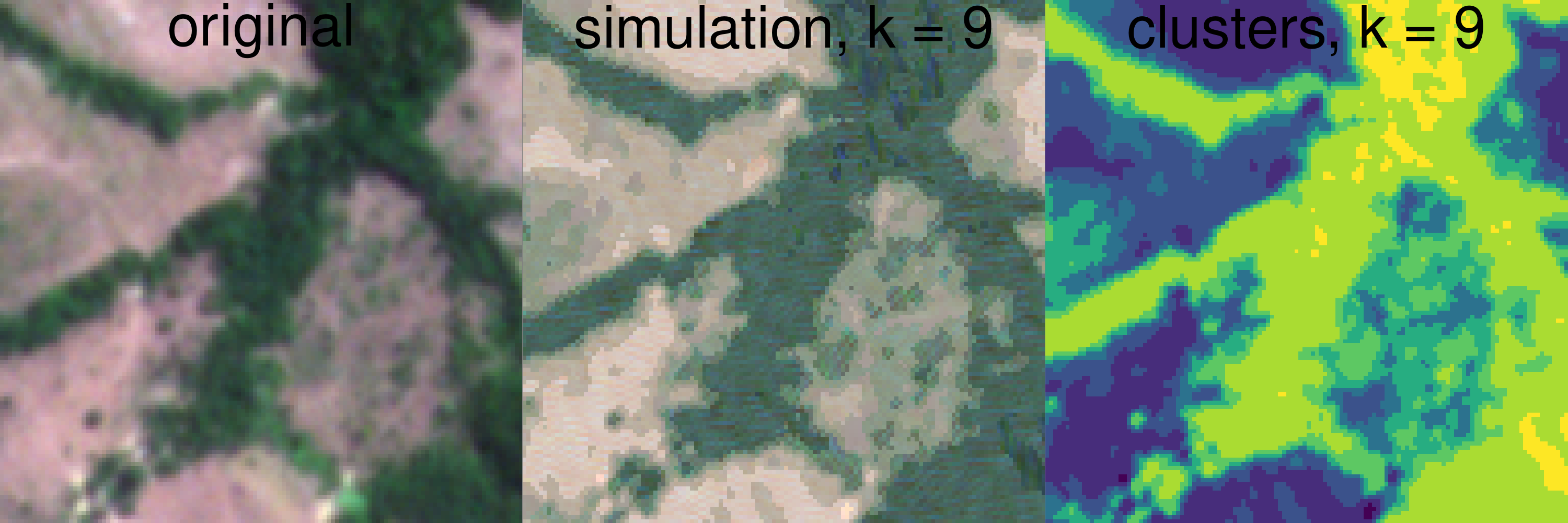}\\
  \centering\includegraphics[width=0.50\linewidth]{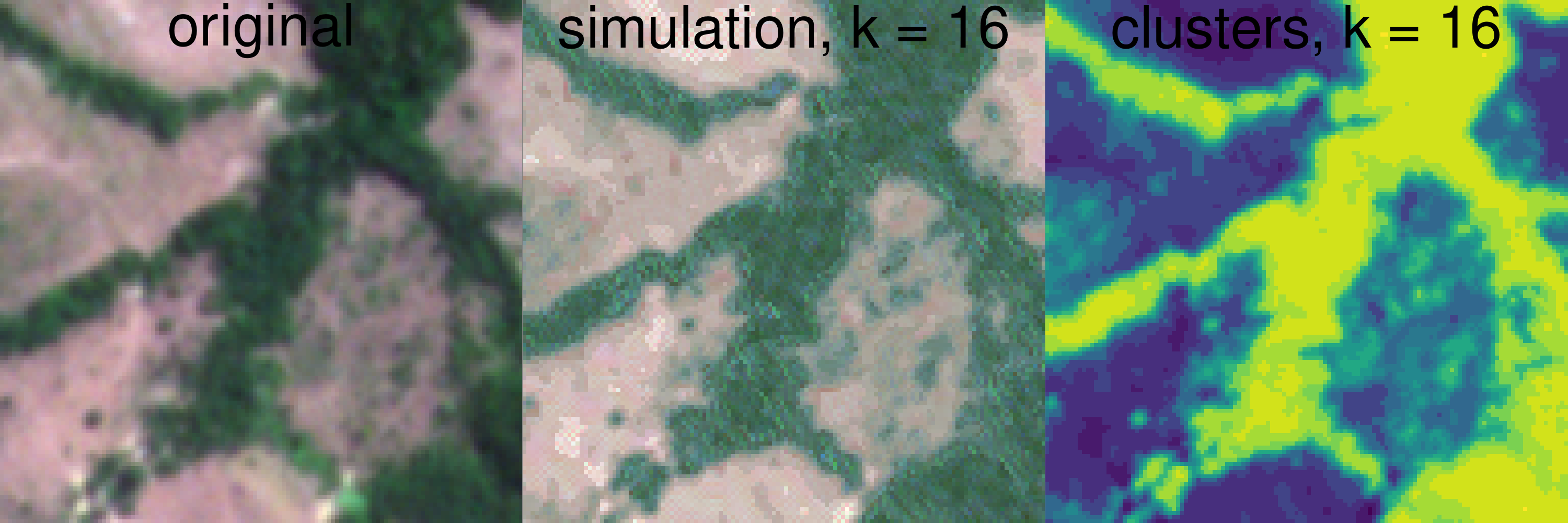}
  \caption{Example of the simulated image and clusters obtained with the k-textures algorithm for a number o classes $k$ = 4, 9 and 16.}
  \label{Fig8}
  \end{figure}

\begin{figure}[H]
  \centering\includegraphics[width=0.80\linewidth]{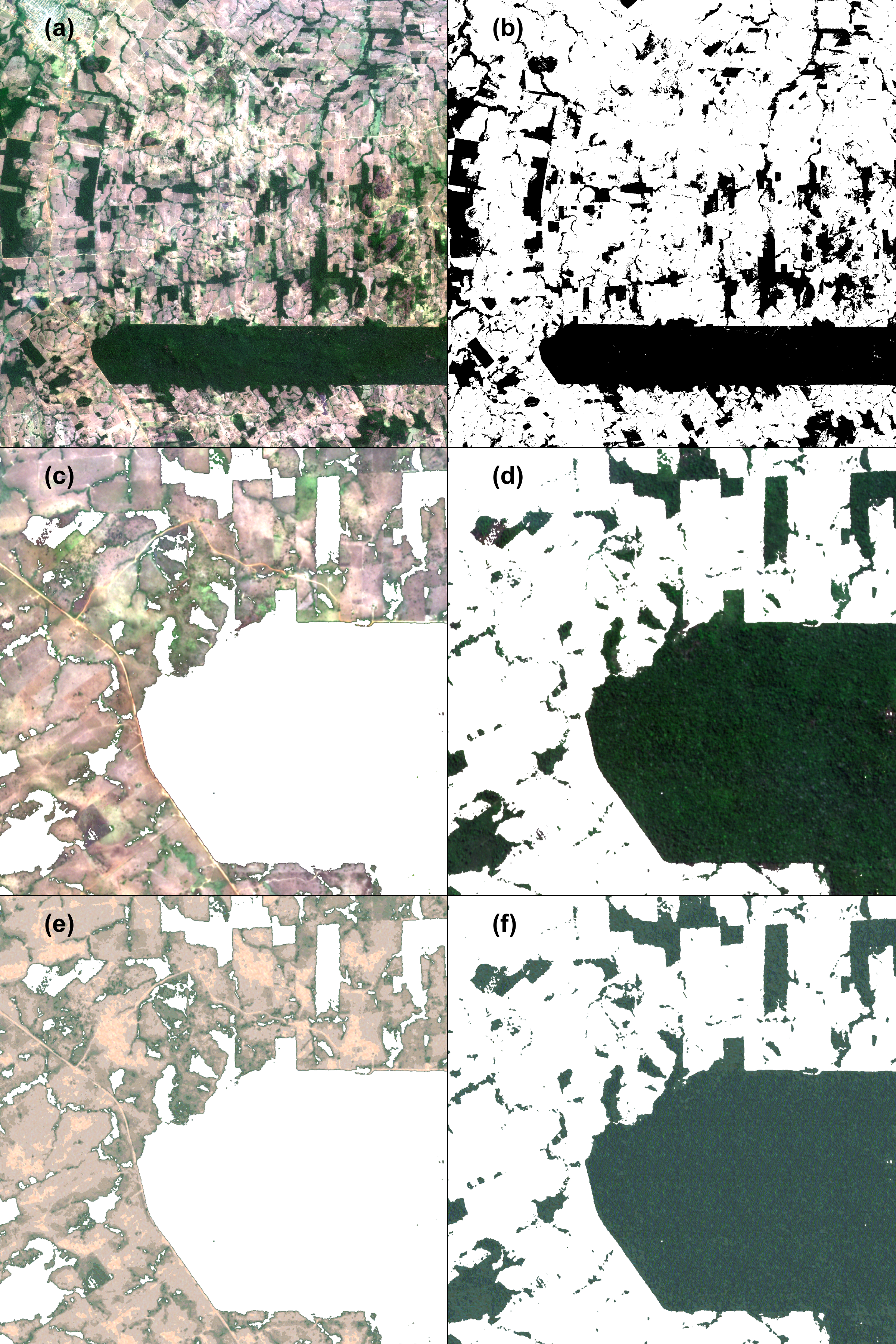}
  \caption{Example of self-segmentation of a Planet image using the k-textures model and $k$ = 9 classes. Original Planet image (a), for visualization only RGB bands are used. Forest  mask produced with k-textures self segmentation (b) where clusters 8 and 9 have been interpreted by the user to be forest and other cluster to represent other land used. Planet image with colors channels set to [255,255,255] for the pixels in the forest mask (c). Planet image with colors channels set to [255,255,255] for the pixels outside the forest mask (d). Simulated Planet image from the k-textures model with colors channels set to [255,255,255] for the pixels in the forest mask (e) and Simulated Planet image from the k-textures model with colors channels set to [255,255,255] for the pixels outside the forest mask (f).}
  \label{Fig10}
  \end{figure}

 In this section, the objective is to self-segment Planet satellite image with the k-textures model to produce highly accurate training sample that could be used to train supervised deep learning model for tropical forest mapping. Just like in k-means, the number of clusters in k-textures should be chosen manually. Finding indices to help determine the best number of $k$ classes for the k-textures model will be addressed in future works. For instance, the user has to visually check the original versus predicted image and mask (for example Fig. \ref{Fig8}) or comparing with other data sets to determine the k-number of classes that seems the best for the segmentation problem, such as the Mapbiomas land cover data. Specifically, for forests, we are searching for classification (i) with textures that reproduce the texture of forest, (ii) without much inclusion of other clusters and (iii) that does not have too much clusters on the pixel representing border or transition classes. With $k$ = 4, Fig. \ref{Fig8}, the model included too many pixels in the cluster corresponding to forest texture. For $k$ = 16, the segmentation does not seem to improve much when looking at the simulated image, in comparison to $k$ = 9, and it seems there is too many clusters inside the forest. However, it can be also noted that the model finds a good texture for the forest and keep this solution with one texture while it had a lot of available classes. For forest segmentation, it can be observed that a k-textures model with $k$ = 9 clusters already provides a good texture representation of the forest,  Fig. \ref{Fig8}. Furthermore, a total of 93.42\% of the pixels classified as forest by Mapbiomas are in the the k-textures clusters 8 and 9, Table \ref{ktkmk9}. Consequently, these two clusters, 8 and 9, were merged and used to make the forest mask, Fig. \ref{Fig10}b.

 The forest mask resulting of the self-segmentation, Fig. \ref{Fig10}b, can be used directly to train a supervised model, such as a U-net model \cite{Ronneberger2015,Wagner2019}. The first advantage is that it gives a class per pixel, consequently it might be more accurate spatially than a manually drawn sample, which generally consist in manual-made polygons that are therefore rasterized which led to large inaccuracy on the pixels at the border of classes. The manual equivalent of our model would be to determine the belonging of a class for each pixel individually. It also has the advantages of the CNNs: the segmentations are fast and consistent \citep{BRODRICK2019734,KATTENBORN202124}, and the pixel are classified using a large number of features, not only the values in the color channels. For example, the original and the simulated images are compared with the 131072 features from the second to last layer of the VGG16 model ($block5\_conv3$ of size 16 $\times$ 16 $\times$ 512) and the 131072 values of the central layer before the decoder of our custom VAE (size 16 $\times$ 16 $\times$ 512). These features are extracted at the end of the encoder for both VGG16 and the VAE model, and, consequently, they represent high level of abstraction. To manually produce the mask given in Fig \ref{Fig10}b would also be extremely time consuming due to the large number of forest patches and likely not as accurate because of the large number of features used in the model which humans are unlikely to capture consistently. The main limitation of the k-textures is that it cannot consider variations that comes from atmospheric condition and shade and it can work properly only if there are little to no variation in illumination inside a class. The model does not account for variation of texture between image patches, for example, the same texture of 128 $\times$ 128 is used for all the patches with forested pixels. To improve this in the future, the model could be adapted to account for variability of the atmospheric condition or shade, for example, the simulated image could be the sum of the texture $\times$ classes added or multiplied to a layer of atmospheric/illumination . 
 
 Comparing the colors and textures of the original image, Fig.  \ref{Fig10}c-d, and the simulated image, Fig. \ref{Fig10}e-f, it can be observed that the model enables to simulate colors and textures that appear visually, to some extents, close to the original image. Note that this is a sub image and the texture of the simulated forest is the same used in the entire image. For the non-forest classes, the model tends to ignore the roads which are of higher hue in the original image. The colors and pattern of roads are smaller, only few pixels, in relation to the size of the texture patch (128 $\times$ 128), so this could be why the model cannot reproduce this spatial pattern. For the forest, the simulated texture is comparable to the real forest textures, with shade and some brighter pixels that are the crowns of the trees. In the \ref{Fig10}f, the pattern of the (128 $\times$ 128) textures patches can be seen. For the production of masks for segmentation, even the k-textures algorithm can produce extremely accurate segmentations, we recommend that the user check visually and correct the mask when necessary. For example for forest, the model is not expected to account for large variations of feature, for example, pink or yellow flowering, clouds and dark shade.

\section{Conclusions}

In this work, we present the k-textures algorithm, a self-supervised algorithm that provides a per pixel self segmentation of an image according to $k$ number of classes. The model uses discrete search with gradient descent, that is, stable binary values are generated and used as weights inside the algorithm. The model remains continuous and fully differentiable by using methods involving the hard sigmoid function and Gaussian noise. To our knowledge, this is the first time stable binary masks are produced and used as weights inside a classical CNN to produce the classes. By contrast to current clustering method using deep learning that all belong to soft clustering, our model returns hard clusters: each pixel has only one class. These stable binary masks enable us to generate a texture per class inside the k-textures architecture. Our model is better than the k-means algorithm for self segmentation because (i) it resolves the same problem, that is, hard clustering of an image in $k$ classes but without assuming a particular distribution of the values inside the clusters and (ii) it uses the capacity of the CNNs, mainly different level of abstraction and spatial context. The k-textures model generates complex textures for each class while the k-means relies only on individual pixel values in the color's channels. Consequently, the k-textures produces textures more similar to the original image. Additionally, the k-textures model sorts the textures along one dimension based on their characteristics. This is likely because of the architecture of the $\sigma_{CNN}$ model used to generate the binary masks and because it helps to decrease the loss. Finally, the hard clustering classes returned by our model can be used for self segmentation in order to produce training sample of objects with particular texture, such as shown for dense tropical forest cover in a Planet image.

\section*{Acknowledgments}
Part of this work was carried out at the Jet Propulsion Laboratory, California Institute of Technology, under a contract with the National Aeronautics and Space Administration (NASA).




\bibliographystyle{unsrtnat}
\bibliography{template.bib}

\end{document}